\documentclass[journal]{IEEEtran}
\usepackage{url}  %Required
\usepackage{graphicx}  %Required

\usepackage{amsthm}
\usepackage{amsmath}
\usepackage{subfig}
\usepackage{algorithm}
\usepackage{algorithmicx}
\usepackage{algpseudocode}
\usepackage{enumerate}
\usepackage{multirow}

\newtheorem*{definition}{Definition}

\usepackage{cite}

\usepackage{mathrsfs}  
\usepackage{bm}
\usepackage{amsfonts,amssymb}
\ifCLASSINFOpdf
\else
\fi
\begin{document}

\title{Explicit and Implicit Pattern Relation Analysis for Discovering Actionable Negative Sequences}

\author{Wei~Wang and %~\IEEEmembership{Member,~IEEE,}
        Longbing~Cao% and Vipin Kumar%,~\IEEEmembership{Senior Member,~IEEE}% <-this % stops a space
        
\thanks{
W. Wang and L. Cao (Corresponding author) are with the University of Technology Sydney, Australia. %, V. Kumar is with the University of Minnesota. 
Email: longbing.cao@uts.edu.au.}
}


\markboth{Preprint}
%\markboth{IEEE Transactions on Knowledge and Data Engineering}
{Shell \MakeLowercase{\textit{et al.}}: Bare Demo of IEEEtran.cls for IEEE Journals}



\maketitle

\begin{abstract}
Real-life events, behaviors and interactions produce sequential data. An important but rarely explored problem is to analyze those nonoccurring (also called negative) yet important sequences, forming \textit{negative sequence analysis} (NSA). A typical NSA area is to discover negative sequential patterns (NSPs) consisting of important non-occurring and occurring elements and patterns. The limited existing work on NSP mining relies on frequentist and downward closure property-based pattern selection, producing large and highly redundant NSPs, nonactionable for business decision-making. This work makes the first attempt for actionable NSP discovery. It builds an NSP graph representation, quantify both explicit occurrence and implicit non-occurrence-based element and pattern relations, and then discover significant, diverse and informative NSPs in the NSP graph to represent the entire NSP set for discovering actionable NSPs. A DPP-based NSP representation and actionable NSP discovery method EINSP introduces novel and significant contributions for NSA and sequence analysis: (1) it represents NSPs by a determinantal point process (DPP) based graph; (2) it quantifies actionable NSPs in terms of their statistical significance, diversity, and strength of explicit/implicit element/pattern relations; and (3) it models and measures both explicit and implicit element/pattern relations in the DPP-based NSP graph to represent direct and indirect couplings between NSP items, elements and patterns. We substantially analyze the effectiveness of EINSP in terms of various theoretical and empirical aspects including complexity, item/pattern coverage, pattern size and diversity, implicit pattern relation strength, and data factors. 
\end{abstract}

\begin{IEEEkeywords}
Negative Sequence Analysis, Negative Sequential Pattern, Pattern Relation Analysis, Pattern Mining, Determinantal Point Process, Nonoccurring Behavior Analysis, Explicit Relation, Implicit Relation, Actionable Pattern Discovery.
\end{IEEEkeywords}

% \fi
\IEEEpeerreviewmaketitle

\section{Introduction}  
\label{secintro}
\subsection{Nonoccurring/Negative Sequence Analysis}

\IEEEPARstart{N}{EGATIVE} sequence analysis (NSA) \cite{WangC19} is a typical method for nonoccurring behavior analytics \cite{nob}. NSA aims to discover interesting negative sequences consisting of negative (non-occurring) and positive (occurring) elements in a sequence. A typical research area in NSA is to discover negative sequential patterns (NSP) \cite{e-NSP2,NSPM,PNSP} that consist of frequent negative sequences. Each \textit{element} may consist of one to multiple \textit{items}, e.g., behaviors. An NSP consists of several negative elements, which may also have positive ones as well. Here, \textit{negative} denoted by symbol `$\urcorner$' means \textit{non-occurring}, e.g., the absence (non-occurrence) of a behavior (an item) or behavior sequence (forming an element or itemset) that is important yet undeclared or missing. 

Examples of negative items are hiding external income (thus undeclared in transactions so as to obtaining government low-income allowance) and missing an important appointment (such as a medical treatment which may incur serious health issues). For example, in COVID-19, an NSP could be $p = <Attending-party, (\urcorner(Wearing-mask), \urcorner(Social-distancing))> \longrightarrow Infected$, where \textit{Attending-party}, $\urcorner(Wearing-mask)$ and $\urcorner(Social-distancing)$ are three items with the first one as a \textit{positive item} and the latter two as \textit{negative items}, $(\urcorner(Wearing-mask), \urcorner(Social-distancing)) = \urcorner(Wearing-mask, Social-distancing)$ as an \textit{negative element}, $<Attending-party, (\urcorner(Wearing-mask), \urcorner(Social-distancing))>$ as a \textit{negative sequence}. We use `,' to separate elements, where an element consisting of more than one item is a compound element, while an element with only one item is a single item element. Pattern $p$ is an impact-targeted negative sequential pattern with an impact label $Infected$ \cite{CaoZZ08}. Note, often impact labels are not necessarily included in NSP and positive patterns. Pattern $p$ indicates that one who attends parties without wearing masks and maintaining social distancing has a high probability of being infected by COVID-19. This NSP reiterates the importance of wearing masks and social distancing to avoid COVID-19 infection, where infection  likely occurs in the absence of two strongly encouraged precautionary behaviors \textit{wearing-mask} and \textit{social-distancing}.  

NSPs are usually more informative and useful than positive sequential patterns (PSPs) (e.g., \cite{Eirini14,AprioriAll,Kiran20}) as they can disclose non-occurring but important behaviors \cite{nob}. NSP mining can disclose the unique NSP value of modeling sequential non-occurring entities, behaviors, and events, including understanding negative sequences \cite{PNSP}, non-occurring yet important behaviors \cite{nob}, and complex behavioral relationships \cite{beg2009fixed,Logic2012coupled,CaoZZ08}. They have been applied to business problems such as fraudulent health insurance claim detection \cite{e-NSP2}, missing medical treatment detection \cite{Neg-GSP,GA-NSP,research}, debt detection in taxation and social welfare services \cite{zhao2009debt,zhao2009mining,CaoZZ08,ZhengWLCCB16}, over-service detection \cite{CaoZZ08,zhao2009mining}, and factors associated with poor academic performance \cite{jiang2018campus}.

\subsection{Gap Analysis and Target Problem}
However, both NSA and NSP discovery involve significant theoretical and practical challenges. Examples are modeling negative elements with diverse formats and combinations; negative patterns mixed with negative items and elements in sophisticated combinations; various negative containment scenarios \cite{e-NSP2}; and different couplings between negative and positive items, elements, and patterns \cite{cao2013combined}. Addressing these learning issues not only requires new theories and tools but also results in high computational cost, large frequent but overlapped findings, and missing significant yet infrequent behaviors, making existing NSA and NSP non-actionable \cite{GA-NSP,e-NSP2,cao2013combined,DongLXW15}. These challenges cannot be directly addressed by existing PSP mining methods \cite{WangC19,e-NSP2}, general sequence analysis methods, and deep neural networks-based sequential modeling \cite{16_mahasseni2017unsupervised}. They only model the occurring items in a sequence but do not handle sophisticated structures and relations in the NSP format, combination, or containment. Very limited theoretical and algorithmic work is available to effectively and efficiently address these learning challenges. 

To address the aforementioned issues, we propose to discover actionable NSPs that (1) are representative, highly probable and diverse to represent the whole original NSP collection; (2) filter highly similar and redundant patterns for low computational complexity and high efficiency \cite{e-NSP2}; and (3) are discriminative and informative for suggesting decision-making actions \cite{cao2013combined}. We call this \textit{actionable NSP discovery}, motivating this work. 
%We aim to discover NSPs which are not just highly frequent (as in the PSP and existing NSP methods) but also informative to decision making. 
Actionable NSP discovery involves several essential but difficult issues: (1) analyzing the relations between NSP elements and patterns, i.e., pattern relation analysis \cite{cao2013combined} to understand how NSP patterns interact with each other, which has rarely been explored; (2) efficiently representing the explicit (e.g., occurrence-based) and implicit (e.g., hidden and indirect) couplings between positive and negative items, elements and their combinations \cite{WangC19}, unexplored in NSA; (3) efficiently handling various sequential combinatorial issues related to item and element combinations, formatting, and containment, very challenging to achieve; and (4) quantifying the optimal selection criteria on each subset candidate in the whole NSP collection (with an emphasis on quality rather than frequency). No existing research on NSA, behavior computing, sequence analysis, and deep sequential modeling addresses these challenges. SAPNSP \cite{DongLXW15} appears to be the only one somehow relevant to representative NSP mining. It follows traditional frequentist by applying the \textit{lift}-based \textit{contribution} measure proposed in \cite{zhao2009mining} to evaluate the contribution of each pattern and select those highly significant ones as the top-k recommendations. It, however, completely ignores the complex couplings between elements and between patterns \cite{cao2013combined,Coupling2015cao,LijffijtSKB16} and the statistical significance of such pattern relations in selecting NSPs, thus producing many redundant and similar patterns.

\subsection{Main Design and Contributions}
In actionable NSP discovery, on one hand, since elements in the NSP cohort often form an imbalanced distribution, patterns that satisfy the \textit{frequentist} selection criteria are often similar and indicate less informative knowledge. The discovered NSPs are associated with a low diversity, a low coverage, and often a heavy-headed distribution of pattern subsets, which miss those infrequent but important items and thus are nonactionable \cite{2_kulesza2011learning,13_gillenwater2014expectation}. The frequentist-based NSA methods and pattern selection criteria widely used in PSP mining and sequence analysis filter patterns with rarely observed items due to their sparsity and the biased selection criteria. In reality, some of the long-tailed items may stand for rare but vital behaviors and are critical in specific situations \cite{nob}, such as suspicious health claims in fraud detection, fault maintenance in system diagnosis, and missing treatments in medical services. On the other hand, the NSPs selected by frequency-based measures and per the downward closure property are often short in size and duplicate, making the discovered pattern subsets less capable of disclosing long-range behaviors and informing decision-making actions. This indicates the importance of measuring \textit{NSP actionability}, such as quality, frequency, probability, diversity and informativeness \cite{cao2013combined} in actionable NSP discovery to serve multiple purposes. One is to select high-quality patterns that carry important multi-aspect information about the representative characteristics in the data. The other is to discover diverse-sized patterns as a group that ensure non-repetitive but informative subset representatives. Intuitively, \textit{pattern diversity} implies a repulsive interaction and negative dependency between NSPs so that similar ones less likely co-occur \cite{pemantle2000towards,borcea2009negative}. In the end, those patterns of low frequency but with more informative knowledge are retained and informative. 

Accordingly, we apply determinantal point processes (DPPs) to representing NSPs as a DPP-based NSP graph, model the explicit (directly linked) and implicit (indirectly linked) pattern relations \cite{cao2013combined}, and then discover actionable NSP subsets. DPP as an efficient probabilistic model captures negative correlations for subset selection by keeping the diversity in the subset %to give the likelihood of selecting a subset
\cite{0_kulesza2012determinantal,01_borodin2009determinantal,8_affandi2014learning,10_blaszczyszyn2018determinantal,12_mariet2015fixed}. However, DPP cannot discover actionable NSPs. 
Building on our work on implicit coupling learning \cite{Coupling2015cao} and its application in inferring indirect rules \cite{wang2017inferring}, this work introduces non-co-occurrence-based implicit relations between NSP itemsets in the DPP-based NSP graph and captures the implicit couplings between two NSPs (or NSP itemsets, e.g., paths $p_1$ and $p_2$) conditioned on other NSP elements or itemsets (e.g., path $p_3$). This method is even more effective in discovering implicitly coupled elements and patterns,  being more informative and discovering even unexpected hidden knowledge \cite{beg2009fixed,cao2013combined}. 

Accordingly, an \textit{explicit and implicit element/pattern relations-based actionable NSP discovery}  method EINSP conducts the DPP-based actionable NSP discovery. 
%EINSP consists of four steps: constructing DPP-based NSP graph, modelling explicit relations between itemsets, modelling implicit relations between itemsets, and selecting the NSPs based on the overall relations combining explicit and implicit relations.
%It first converts NSP discovery to DPP-based NSP subgraph selection, and a path in the  DPP-based NSP graph corresponds to an NSP. The DPP-based NSP graph models both co-occurrence-based explicit relations and non-occurrence-based implicit relations between NSP elements and patterns. EINSP then discovers a representative subset of high-probability and diverse NSPs in the DPP-based NSP graph. 
%Our proposed DPP-based representative NSP discovery method makes the following original contributions.
\begin{itemize}
    \item This is the first work on designing a DPP-based graph representation of NSPs and discovering high-quality NSPs that are representative and of high probability and high diversity in the NSP graph. The problem of representative NSP discovery is converted to a probabilistic subset selection problem in a DPP graph, which takes advantage of the probabilistic DPP theories and graph representation strength in subset selection with diversity. We derive the DPP-based theory for representative NSP discovery and design EINSP to select a subset of informative and diverse NSPs using DPP-based subset selection. 
    \item EINSP captures rich element interactions and pattern relations in NSPs, which are rarely explored in existing NSA and sequence analysis. The co-occurring and non-occurring element and pattern relations in NSPs are modeled in terms of the direct and indirect DPP-based node/edge dependencies, which characterizes the probability and diversity of each NSP pattern in the NSP collection in terms of both explicit element/pattern co-occurrences and implicit non-occurrences conditional on third parties. EINSP integrates both explicit and implicit element/pattern relations in the DPP-based NSP graph and effectively sample those highly explicitly and implicitly coupled NSPs as a representative subset of high-probability and diverse NSP patterns.
\end{itemize}

We verify EINSP on six real datasets and 17 synthetic datasets against baseline methods in terms of sequence and item coverage, average pattern size, average implicit relation strength, and sensitivity to various data factors including scalability. The substantial empirical analysis demonstrates that EINSP achieves significant performance in discovering representative NSPs. It opens new opportunities for efficient NSA through DPP-based representation and learning.

\section{Background and Related Work} 
\label{secRW}

\subsection{NSA, NSP Mining, and Nonoccurring Behavior Analysis}
\textit{Sequence analysis}\footnote{Note: here we do not involve the general sequence analysis such as deep sequential modeling and other topics including time series analysis.} typically discovers patternable combinations of items/elements in sequences, e.g., frequent sequential patterns \cite{AprioriAll,LijffijtSKB16}, which is also termed \textit{(positive) sequence analysis} (PSA) \cite{Eirini14,4_yuan2016discovering,ZhouCG15,9_li2019short,HassaniTCS19,RallaRM19} in particular \textit{positive sequential pattern} (PSP) mining \cite{AprioriAll,Kiran20}. PSA and PSP mining identify frequently co-occurring itemsets. In contrast, NSA and NSP mining \cite{e-NSP2,WangC19} discover non-occurring but interesting items/elements and itemsets. Typically, PSA and NSA focus on selecting frequent patterns. 

While PSA and in particular PSP mining have been intensively explored, much less effort has been dedicated to developing basic theories and efficient computational tools for NSA and NSP mining in its over a decade history \cite{WangC19}. This is due to the intricate characteristics and challenges of negative items, elements and sequences \cite{e-NSP2} and the fact that the approaches and tools developed for PSA and PSP mining cannot be directly transferred or lightly adapted to NSA and NSP mining since they hold highly different problem settings and complexities. Existing work on NSA and NSP mining involve various aspects of constraints to restrict the settings of negative candidates and NSPs and to reduce or control the high combinatorial challenge and computational complexity \cite{e-NSP2,WangC19}. Examples are element size constraints, element format constraints, pattern structure constraints, and containment constraints. Note, even the same constraint settings \cite{WangC19,SmedtDW20} could incur different challenges and solutions in PSP and NSP mining. Further, NSA only forms a small set of the even broader and more challenging problem - \textit{nonoccurring behavior analysis} (NBA) \cite{nob}. NBA involves many rarely considered problems, settings and opportunities of behaviors that are important but have not yet happened (non-occurred) or observed (unobserved, or hidden).

Major approaches for NSA and NSP mining can be categorized into two: (1) the \textit{frequentist-based NSA} that develops frequency-based statistics including \textit{support} and \textit{confidence} of negative candidates and NSPs and for their selection \cite{Neg-GSP,e-NSP2,WangC19,GuyetQ20}; and (2) the \textit{set theory-based NSA} and its landmark algorithm e-NSP proposed in \cite{e-NSP2} by converting NSA to PSP and then conducting PSP-based NSA. The former approach transfers the theories of PSA and PSP mining to NSA and NSP mining but cannot handle the fundamental challenges of NSA due to their intrinsic differences. As a result, very limited progress has been made so far. The latter approach opens a fundamental new direction to address the high combinatorial challenge and computational complexity in NSA, leading to various recent follow-ups \cite{GongXDL17,DongGC18,DongQLCX19,DongGC20,HuangWJ20}. However, their methodological dependence on frequentist statistics incorporates various constraints (e.g., FRI in \cite{HuangWJ20}) rules out infrequent items and elements, and none of them involves relation modeling of elements and patterns as in our work.

In contrast, this paper opens a new direction for NSA post processing. It avoids the frequentist-oriented combinatorial challenge and computational complexity by converting the NSA problem to subgraph selection after designing a DPP-based graph representation of NSPs. This work is thus highly fundamental and promising for addressing some of the critical challenges in NSA.

\subsection{Actionable NSP Discovery}
NSA and NSP mining still face some critical theoretical and computational challenges. One is to develop general representations of negative sequences and their various settings (e.g., on format, type, containment) and constraints on the combinations of items, elements, and sequences. Another is the evaluation and selection of quality NSPs (here `quality' may refer to aspects of significance, representativeness, novelty, coverage, and actionability). The last example is to analyze explicit and implicit coupling relationships within sequences and between sequences \cite{cao2013combined,Eirini14,Coupling2015cao,LijffijtSKB16}, e.g., pair patterns, and cluster patterns \cite{CaoZZ08,cao2013combined}. 

The challenge addressed in this work is on discovering actionable NSPs \cite{cao2013combined,DongLXW15} that are significant, diverse, and informative for suggesting decision-making actions. This is essential and critical to not only discover more actionable NSPs to enhance NSA applicability such as for next-best medical treatments but also to address some fundamental technical issues such as analyzing sophisticated couplings in NSP elements and sequences. Compared with the high degree of attention paid to general NSP discovery (e.g., \cite{NSPM,Neg-GSP,GA-NSP,e-NSP2}), very limited research has been conducted on selecting representative NSPs that are not only statistically significant but also more actionable (e.g., inducing more diversified and novel elements) to decision-making. 

However, to the best of our knowledge, SAPNSP \cite{DongLXW15} is the only method on NSP subset selection, i.e., extracting top-k NSPs from those NSP collection discovered by other NSP miners. SAPNSP cannot discover actionable NSPs due to three main design shortages: 1) it only applies a single measure \textit{contribution} originally proposed in our early work in \cite{zhao2009mining} to selecting patterns of high frequency and high correlation between its prefix and the last element, producing a highly repetitive resultant subset; 2) the \textit{contribution} measure is based on \textit{lift} and the downward closure property, which thus overestimates the quality of short-size patterns but filters long-size patterns with highly frequent and informative elements; 3) SAPNSP does not consider implicit non-occurrence couplings between elements and between NSPs, and thus filters these lowly frequent but informative patterns. Our approach suggests a comprehensive solution without the aforementioned limitations in discovering actionable NSP discovery. In comparison to \textit{contribution}-based filtering, EINSP calculates both co-occurrence and non-co-occurrence-based dependencies between directly and indirectly linked itemsets in the whole NSP cohort. They thus capture much richer interactions than \textit{contribution} between NSP elements and itemsets and are not restricted to the downward closure property-based element overlapping as \textit{contribution}. In addition, although SAPNSP addresses the interestingness of mining NSPs, its application of \textit{contribution}-based filtering is too simple without an appropriate measure of NSP `actionability' \cite{cao2013combined} and any empirical evaluation of the effectiveness as in our work (see more details in Section \ref{secEA} for a comprehensive evaluation of EINSP effectiveness).

\subsection{DPP and DPP-based Subset Selection}
DPPs have shown great strength in subset selection by modeling the probability over all subsets in terms of their quality and diversity. 
DPP has shown promise for video and documentation summarization \cite{2_kulesza2011learning,7_gong2014diverse,15_hong2014improving,16_mahasseni2017unsupervised}, information retrieval \cite{3_gillenwater2012discovering}, recommender systems \cite{6_chen2018fast}, sequence classification \cite{9_li2019short}, and image processing \cite{4_yuan2016discovering, 11_kulesza2011k}. However, DPP has not been used for actionable NSP discovery. On one hand, NSPs are embedded with sequential structures, which cannot be modeled by the point-based DPP methods. So far, structured DPP (SDPP) \cite{1_kulesza2010structured} is the only DPP-based method for distributions over sets of structures and was applied to documentation summarization \cite{3_gillenwater2012discovering}. However, it only models the quality and diversity of structures in terms of a single relation but fails to handle multiple complex relations between entities, which may cause unacceptable information loss \cite{4_yuan2016discovering}. On the other hand, the relations between NSPs are much more complicated than the dependency modeled by the existing DPP-based methods. The method in \cite{3_gillenwater2012discovering} incorporates  cosine similarity into SDPP to model the quality of each entity and quantify the diversity between two entities (an \textit{entity} can be a structure or a component within the structure), which only reveals the co-occurrence relations. 
In addition, \textit{k}-DPP only selects a significant and diverse subset with cardinality $k$ \cite{11_kulesza2011k}. MDPP refined on \textit{k}-DPP selects an optimal fixed-size subset by modeling multiple relations among the entities in the collection \cite{4_yuan2016discovering}. However, neither \textit{k}-DPP nor MDPP are applicable for selecting structural patterns like NSPs. 

Our work represents the first attempt at engaging DPP for NSP mining for NSP quality enhancement and representative subset selection. No research has been reported on effectively representing NSA in terms of DPP. Our proposed EINSP represents the highly probable, diverse and coupled NSPs by DPP, jointly modeling co-occurrences and non-occurrences-based explicit and implicit itemset/pattern relations \cite{cao2013combined} which have rarely been explored but are very important to PSA and NSA research, and selecting representative NSP subsets as the NSA output.

\section{The EINSP Method} 
\label{sec:EINSP}

\subsection{Problem Statement and System Framework}

Let us illustrate the aforementioned motivation of actionable NSP discovery with a more general healthcare analytical case. Assume $p=<a,\urcorner b, c, X>$ is an NSP, where $a$, $b$ and $c$ stand for the codes of the medical services undertaken on a patient, and $X$ is the final disease status of the treatment. $p$ indicates that patients who undertake medical services $a$ and $c$ without treatment $b$ have a high probability of having disease status $X$, which shows the impact of the absence of $b$ on status $X$. Pattern $p$ may correspond to a special treatment of a serious but low-chance disease, such as a rare cancer, where treatments $a$, $b$ and $c$ are less likely to co-occur frequently. Such patterns may not be selected by existing NSA methods due to its relatively low frequency, even though $p$ may be extremely useful for informing the disease treatment. This example also shows the importance of involving the negative items/elements and their implicit relations with positive items/elements, both of which are usually ignored in PSP and general sequence modeling; the findings are thus more informative by referring to the non-occurring items and infrequent patterns. However, none of the existing DPP-based subset selection methods can model such pattern quality and diversity-related settings and can jointly model positive and negative element/pattern relations. 

\begin{figure*}[!t] 
\centering\includegraphics[width=6.4in]{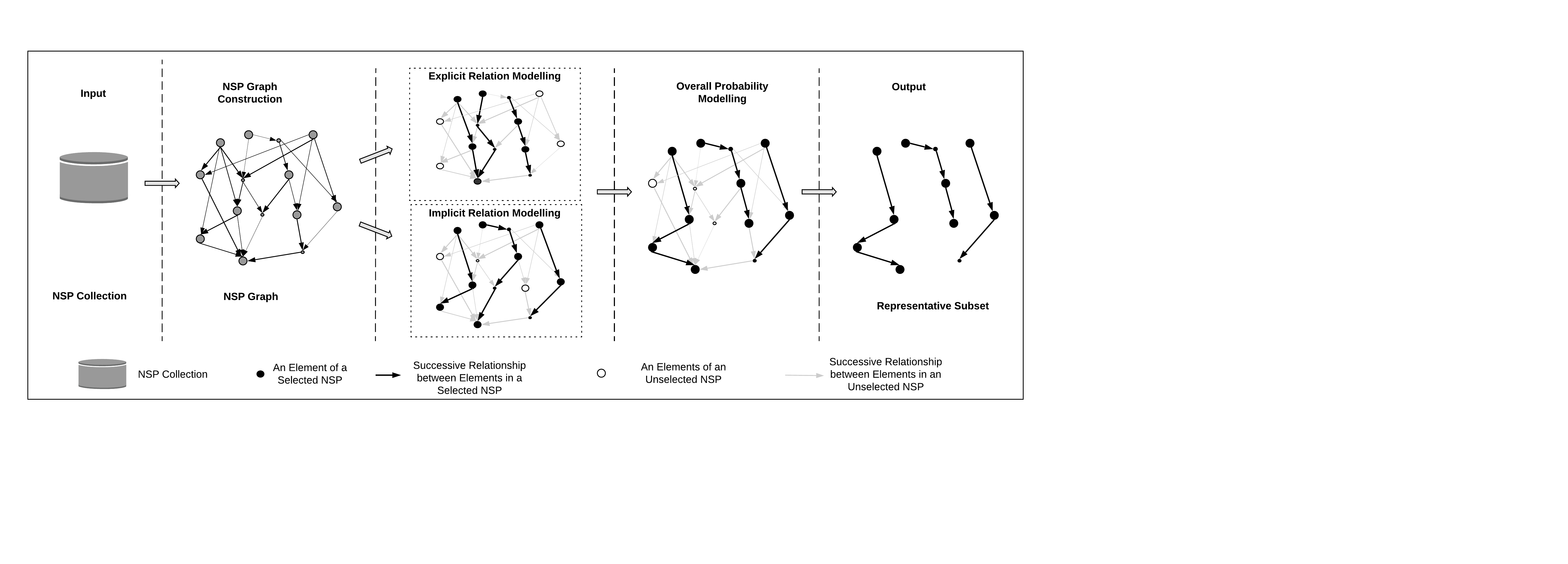} 
\caption{Actionable NSP discovery of high-quality and diverse NSPs by modelling both explicit and implicit element/pattern relations in DPP-based NSP representations.}\label{framework} 
\end{figure*}

To discover actionable NSPs, Fig. \ref{framework} illustrates the working process of the proposed DPP-based actionable NSP discovery EINSP. EINSP consists of four components: NSP graph construction, explicit relation modeling, implicit relation modeling, and overall relation-based NSP subgraph selection. First, the \textit{NSP graph construction} step transforms a collection of NSPs discovered by an NSP miner (e.g., Negative-GSP \cite{Neg-GSP}) into a directed DDP-based graph. Second, the \textit{explicit relation modeling} step models the explicit co-occurring relations between NSPs in the DPP graph and computes the probability of selecting an NSP subset with such explicit relations. Third, the \textit{implicit relation modeling} step models the implicit non-co-occurring relations between NSPs with regard to another NSPs in the DPP graph and computes the probability of selecting an NSP subset with such implicit relations. Lastly, the \textit{overall relation-based selection} computes the overall probability of selecting an NSP subset by integrating both explicit relations-based probability and implicit relations-based probability. A representative NSP subset is selected from the DPP-based NSP graph per the overall probability and selection criteria. EINSP selects $k$-size informative and diverse subsets from the original NSP cohort in the DPP graph.

Accordingly, let us assume $\mathcal{Y}$ be the entire NSP pattern set, $Y_i \in \mathcal{Y}$ refers to the $i-th$ NSP, and $y_i^j$ refers to the $j-th$ element in pattern $Y_i$. Correspondingly, their notations in the converted DDP-based graph are as follows: $\mathcal{G}$ refers to the NSP-converted DPP graph corresponding to the pattern set $\mathcal{Y}$; $\bm{Y_i}$ refers to the path in the DPP graph corresponding to NSP $Y_i$ which represents a vectorized representation of the NSP in $\mathcal{G}$; $y_i^j$ refers to the node in graph $\mathcal{G}$ corresponding to the NSP element $y_i^j$, i.e., an element of vector $\bm{Y_i}$.

Further, assume we have a collection of NSPs $\mathcal{Y}=\{Y_1, Y_2, \ldots, Y_N\}$ discovered by any existing NSP miner as the input, where $Y_i = <y_i^1, y_i^2, \ldots, y_i^{n_i}> \in \mathcal{Y}$ is an NSP and $y_i^j \in Y_i$ is an NSP element in the NSP. EINSP first transforms NSPs $\mathcal{Y}$ into a directed DDP-based graph $\mathcal{G}$, which is a powerful and compressed representation of the NSPs. Second, EINSP models the explicit relations between NSPs in the DPP graph and computes the probability $P^{k}_{e}(Y)$ of selecting an NSP subset $Y \in \mathcal{Y}$ in terms of co-occurrences between NSPs in each $k$-subgraph in the DPP graph. Then, EINSP further represents the implicit relations between NSPs in the DPP graph and computes the probability $P^{k}_{i}(Y)$ of selecting the NSP subset $Y$ in terms of their non-occurrences with other NSPs. Further, EINSP calculates the overall probability of the NSP subset $Y$ by integrating both $P^{k}_{e}(Y)$ and $P^{k}_{i}(Y)$. Lastly, EINSP selects the representative $k$-size NSP subsets as the actionable NSPs, which correspond to the highly probable, diverse and explicitly and implicitly coupled NSP subgraphs in the DPP-based NSP graph per a selection criteria over the overall probability $P^{k}(Y)$. In the following, we introduce each of these modules in detail.

\subsection{NSP Graph Construction} \label{GraphCon}

In general, the NSP collection $\mathcal{Y}$ is always on a large scale and many patterns share identical subsequences (for simplicity and consistency with the terms in NSP, \textit{subsequences} are used interchangeable with \textit{itemsets}) due to the element and itemset co-occurrences \cite{e-NSP2}. Hence, the NSP graph construction module converts $\mathcal{Y}$ to a directed graph $\mathcal{G} = (\mathcal{V}, \mathcal{E})$, where the direction refers to the sequential order between elements in NSPs. In $\mathcal{G}$, each node $y \in \mathcal{V}$ corresponds to an element (consisting of all items in the element) of an NSP pattern, and each edge $e \in \mathcal{E}$ stands for a directed linkage from one element to another in an NSP, which reflects the sequential co-occurrences between the elements in the NSP. %sequential order relationship between a pair of elements inside an NSP. 
In this way, each NSP $Y_i \in \mathcal{Y}$ is transformed into a directed path of $\mathcal{G}$. 

In $\mathcal{G}$, we define\footnote{Due to space limitation, we could not illustrate definitions by toy examples. Interested readers may refer to \cite{WangC19,e-NSP2} for systematic definitions and examples for NSA.} the \textit{explicit element quality}. It quantifies the quality (significance) of each NSP element (i.e., a node significance in the graph) in graph $\mathcal{G}$. We further define the \textit{explicit element pair quality} for each element pair in an NSP (i.e., an edge in $\mathcal{G}$) in terms of the explicit co-occurrence relations between NSP elements. 
\begin{definition}[Explicit Element Quality]
    Given a path $\bm{Y_i} = <y_i^1, y_i^2, \ldots, y_i^{n_i}>$ in $\mathcal{G}$ which corresponds to an NSP pattern $Y_i$ in the pattern set $\mathcal{Y}$, the \textit{explicit element quality} $q_e(y_i^j)$ measures the significance of pattern element $y_i^j$ in $\mathcal{G}$, and the \textit{explicit element pair quality} $q_e(y_i^j, y_i^{j+1})$ measures the co-occurrences-based significance of the pattern element pair $<y_i^j, y_i^{j+1}>$ in $\mathcal{G}$.
\end{definition}

In addition, we define the \textit{explicit element diversity feature vector} for each NSP element in $\mathcal{G}$.
\begin{definition}[Explicit Element Diversity Feature Vector]
    Given a node $y_i^j \in \bm{Y_i}$ in $\mathcal{G}$ which corresponds to an element $y_i^j$ of NSP $Y_i$, its \textit{explicit element diversity feature vector} $\phi_e(y_i^j|E)$ %$\phi_e(y_i^j) \in R^{\vert E \vert}$
    measures the explicit co-occurrence diversity between NSP element $y_i^j$ and other co-occurring NSP elements in $\mathcal{G}$, where $E$ is the set of potential elements of NSPs and $\vert E \vert$ is the size of all elements in $E$.
\end{definition}

Lastly, we define the \textit{implicit pattern quality} and the \textit{implicit pattern diversity feature vector} for each NSP in $\mathcal{G}$.
\begin{definition}[Implicit Pattern Quality]
    Given a path $\bm{Y_i}$ in graph $\mathcal{G}$ which corresponds to an NSP $Y_i$ in the pattern set $\mathcal{Y}$, the \textit{implicit pattern quality} $q_i(\bm{Y_i})$ measures the quality of the path in $\mathcal{G}$ in terms of implicit relations between paths, i.e., the non-occurrence level between $Y_i$ and other NSPs in the DPP graph.
\end{definition}
 
\begin{definition}[Implicit Pattern Diversity Feature Vector]
    Given a path $\bm{Y_i}$ in graph $\mathcal{G}$ corresponding to an NSP $Y_i$ in the pattern set $\mathcal{Y}$, the \textit{implicit pattern diversity feature vector} $\phi_i(\bm{Y_i})$ measures the diversity of the path in $\mathcal{G}$ in terms of the non-occurrence relations between $\bm{Y_i}$ and other paths, i.e., the non-occurrence level between pattern $Y_i$ and other NSPs in the DPP graph.
\end{definition}

The above basic concepts form the base for the following formalization of DPP-based NSP representation and learning. 

\subsection{Explicit Relation Modelling} \label{ExpRelMod}

Since each pattern $Y_i \in \mathcal{Y}$ represents an ordered structure of NSP elements, the occurrence probability (also called \textit{explicit probability}) of an NSP subset $Y$ can be computed by a fixed \textit{k}-size structured determinatal point process (\textit{k}-SDPP) model \cite{1_kulesza2010structured}, which captures the co-occurrence-based relationships between NSP elements and their formation into NSP patterns. The \textit{k}-SDPP model captures the distribution of pattern subset $Y \subseteq \mathcal{Y}$ conditional on the event that  subset $Y$ has cardinality $k$ as follows. 

%\begin{small}
\begin{equation} \label{ExpProb}
    P^k_e(Y) = \frac{det(L^{e}_{Y})}{\sum\nolimits_{\vert Y^{\prime} \vert=k} det(L^{e}_{Y^{\prime}})}
\end{equation}
%\end{small}
%
Here, $L^{e}$ is a positive semidefinite kernel, $L^{e}_{Y}$ represents the restriction on $L^{e}$ to entries indexed by NSP subset $Y$ \cite{11_kulesza2011k}, i.e., $L^{e}_{Y} \equiv [L^{e}_{ij}]_{\bm{Y_i}, \bm{Y_j} \in \mathcal{Y}}$, and $det(L^{e}_{Y})$ denotes the determinant of $L^{e}_{Y}$. 

Following the design in \cite{0_kulesza2012determinantal}, kernel $L^{e}$ can be rewritten as a Gram matrix: $L^{e}={B^{e}}^{T}B^{e}$. $B^{e} \in R^{\vert E \vert \times \vert \mathcal{Y} \vert}$ is the matrix where each column is a feature vector  \cite{0_kulesza2012determinantal}, describing the corresponding pattern in the NSP collection $\mathcal{Y}$. Column $B^{e}_{i}$ can be factorized as the production of an \textit{explicit pattern quality} score $q_e(\bm{Y_i}) \in R^{+}$ and a normalized \textit{explicit pattern diversity feature vector} $\phi_e(\bm{Y_i})$, %\in R^{\vert E \vert}$, 
$\lVert \phi_e(\bm{Y_i}) \rVert=1$. Here, $q_e(\bm{Y_i}) \in R^{+}$ can be viewed as a non-negative measure of the explicit significance of pattern $Y_i$, and $\phi_e(\bm{Y_i})^{T}\phi_e(\bm{Y_k}) \in [-1,1]$ as a signed measure of explicit similarity (diversity) between patterns $Y_i$ and $Y_k$. 

Accordingly, the entries of kernel $L^{e}$ can be further decomposed as follows:
\begin{equation} \label{ProDec}
    L^{e}_{ik}=q_e(\bm{Y_i})\phi_e(\bm{Y_i})^{T}\phi_e(\bm{Y_k})q_e(\bm{Y_k})
    %L^{e}_{ij}=B_{e}^{T}B_{e}=q_e(\bm{y_i})\phi_e(\bm{y_i})^{T}\phi_e(\bm{y_j})q_e(\bm{y_j})
    %P^k_e(Y) = \frac{det()}{\sum\nolimits_{\vert Y^{\prime} \vert=k} det(L^{e}_{Y^{\prime}})}
\end{equation}
Eq. (\ref{ProDec}) gives rise to a distribution \cite{2_kulesza2011learning}, which places a higher weight on the NSP subsets that are composed of higher quality  and more diverse NSP patterns. 

To efficiently define a DPP over the sequential NSP patterns, the explicit pattern quality score $q_e(\bm{Y_i})$ is multiplicatively decomposed as follows. We use a log-linear model to decompose it, which depends on the explicit quality of each element $q_e(y_i^j)$ and its element pair $q_e(y_i^j, y_i^{j+1})$ for two elements $y_i^j$ and $y_i^{j+1}$.

\begin{small}
\begin{equation} 
\begin{split} \label{q_e} %\label{ExpQua}
    q_e(\bm{Y_i})&=exp(\sum\nolimits_{j=1}^{n_i} q_e(y_i^j) + \sum\nolimits_{j=1}^{n_i-1} q_e(y_i^j,y_i^{j+1})) \\
    &=\prod\nolimits_{j=1}^{n_i} exp(q_e(y_i^j)) \times \prod\nolimits_{j=1}^{n_i-1} exp(q_e(y_i^j,y_i^{j+1}))
\end{split}
\end{equation}
\end{small}

As  NSP discovery is built on  frequentist statistics, we specify the \textit{explicit quality of a pattern feature} $d$ (corresponding to the element and element pair in the feature set) in terms of element and element pair frequencies in NSPs, i.e., $q_e(\centerdot) \equiv \frac{\vert \{ s_{d} \vert s_{d} \in D \wedge <\centerdot> \subseteq s_{d} \} \vert}{\vert D \vert}$.  $D$ denotes the pattern set and $s_{d}$ is an NSP sequence from dataset $D$. Similarly, the normalized \textit{explicit pattern diversity feature vector} $\phi_e(\bm{Y_i})$  of an NSP pattern $Y_i$ is decomposed additively over the elements $y_i^j$:

\begin{equation}
    \phi_e(\bm{Y_i}) = Norm(\sum\nolimits_{j=1}^{n_i} \phi_e(y_i^j))
\end{equation}

Here, $Norm(\centerdot)$ guarantees that, for any pair of NSP patterns $\bm{Y_i}$ and $\bm{Y_k}$, $\phi_e(\bm{Y_i})^{T} \phi_e(\bm{Y_k}) \in [-1,1]$ is a signed measure of the similarity between these patterns. The diversity feature $d^{e}_{k}(y_i^j) \in \phi_e(y_i^j)$ is the $k$-th component of vector $\phi_e(y_i^j)$. The $k$-th component is identified by the \textit{explicit normalized element-wise mutual information} (eNEMI) between NSP element $y_i^j$ and other NSP elements $E_k \in E$. eNEMI is adapted from the \textit{normalized point-wise mutual information} (NPMI) \cite{bouma2009normalized}.

\begin{definition}[Explicit Normalized Element-wise Mutual Information]
    The \textit{explicit normalized element-wise mutual information} (eNEMI) $NEMI(y_i^j,E_k)$ measures the ability to capture both linear and non-linear dependencies between element $y_i^j$ and other NSP elements $E_k \in E$, which is defined as follows:
    \begin{small}
    \begin{equation}\label{eq:eNEMI}
        d^{e}_{k}(y_i^j) \equiv eNEMI(y_i^j,E_k) =\frac{h(y_i^j)+h(E_k)-h(y_i^j,E_k)}{h(y_i^j,E_k)} 
    \end{equation}
    \end{small}
\end{definition}
where $h(\centerdot)=-log p(\centerdot)$, $p(y_i^j)$ and $p(E_k)$ are the marginal probabilities of NSP elements $y_i^j$ and $E_k$ in  graph $\mathcal{G}$. $p(y_i^j, E_k)$ is the joint probability. Note that $d^{e}_k(y_i^j) \in [-1,1]$, and some orientation values \cite{bouma2009normalized} are as follows: (1) when NSP elements $y_i^j$ and $E_k$ occur separately but never co-occur, they hold a negative dependence corresponding to non-co-occurrences between NSP elements, $d^{e}_k(y_i^j)=-1$; (2) when they are distributed under independence, $d^{e}_k(y_i^j)=0$ as the numerator is 0; and (3) when they completely co-occur, they hold positive dependence corresponding to the full co-occurrences between NSP elements, $d^{e}_k(y_i^j)=+1$. Hence, $\phi_e(y_i^j)$ serves as an \textit{element dependence measure} of $y_i^j$ w.r.t. all other elements. 

Discovering NSP involves a huge search space and a high computational cost. The size of the NSP collection $\mathcal{Y}$ is always enormous \cite{e-NSP2,PNSP,Neg-GSP}, which makes  kernel $L^{e}$ too large and the inference on $L^{e}$ intractable. Accordingly, a \textit{dual representation} of $L^{e}$, i.e., $C^{e} = {B^{e}}{{B^{e}}^T}$, is constructed to represent the properties carried by kernel $L^{e}$ to enable efficient inference \cite{0_kulesza2012determinantal}. Note that $C^{e} \in R^{\vert E \vert \times \vert E \vert}$ is a symmetric and positive semidefinite matrix, where typically $\vert E \vert \ll \vert \mathcal{Y} \vert$. Hence, $C^{e}$ is always much smaller in scale and less sensitive to a threshold than  kernel $L^{e}$. 

Accordingly, the dual representation $C^{e}$ of NSP $Y_i$ in the NSP set $\mathcal{Y}$ can be factorized as follows through the DPP-based representations:
% $N = \vert \mathcal{Y} \vert$
\begin{equation} \label{DualRepr}
\begin{split}
    %C&= BB^T\\
    C^{e}=\sum\nolimits_{\bm{Y_i} \in \mathcal{Y}} q_{e}^2(\bm{Y_i})\phi_e(\bm{Y_i})\phi_e(\bm{Y_i})^{T}\\
    %&=\sum\nolimits_{\bm{y_i} \in \mathcal{Y}} (\prod\nolimits_{j=1}^{n_i} exp^2(q_e(y_i^j)) \times \prod\nolimits_{j=1}^{n_i-1} exp^2(q_e(y_i^j,y_i^{j+1}))) (Nor(\sum\nolimits_{j=1}^{n_i-1} \phi_e(y_i^j))) (Nor(\sum\nolimits_{j=1}^{n_i-1} \phi_e(y_i^j)))^{T}
\end{split}
\end{equation}
The representation $C^{e}$ can be calculated by a second-order message passing algorithm \cite{li2009first,01_borodin2009determinantal}. %It involves the time complexity $O({\vert E \vert}^2{\vert \mathcal{Y} \vert})$. 

Once $C^{e}$ is computed, it can further be eigen-decomposed in the form of $C^{e}=\sum\nolimits_{n}\lambda^{e}_n v^{e}_n {v^{e}_n}^{T}$ in time $O({\vert E \vert}^{3})$. We then obtain the eigenvalue/eigenvector pairs ${(\lambda_n, v_n)}_{N_v}$ of representation $C^{e}$, where $N_v$ is the number of eigenvalue/eigenvector pairs. As proved in \cite{0_kulesza2012determinantal}, the non-zero eigenvalues of $C^{e}$ and $L^{e}$ are identical. If $v_n$ is the $n$-th eigenvector of $C^{e}$ then ${B^{e}}^{T}v_n$ is the $n$-th eigenvector of $L^{e}$, which share the same eigenvalue $\lambda^{e}_n$. That is, ${(\lambda^{e}_n, {B^{e}}^{T} v^{e}_n)}_{N_v}$ are the corresponding pairs of kernel $L^{e}$, making $L^{e}=\sum\nolimits_{n}\lambda^{e}_n  {({B^{e}}^{T} v^{e}_n)} {({B^{e}}^{T} {v^{e}_n})}^{T}$. 

To efficiently select a subset of $k$ patterns, inspired by the mechanism of \textit{k}-DPP \cite{11_kulesza2011k}, we formalize $P^k_e(Y)$ as follows: 

\begin{equation} \label{ExpProFor}
    P^k_e(Y)= \frac{1}{e_{k,{N_v}}^{e}} \sum\limits_{{\vert J \vert =k} \wedge J \subseteq \{1,2,\ldots,N_v\}} P^{V^{e}_J}(Y) \prod\limits_{n \in J} \lambda^{e}_n
    %P^k_e(Y)= \frac{1}{e_k^{{N_v}}} \sum\limits_{{\vert J \vert =k} \wedge J \subseteq \{1,2,\ldots,N_v\}} P^{V^{e}_J}(Y) \prod\limits_{n \in J} \lambda^{e}_n
\end{equation}
Here, $J$ is the index subset of ${(\lambda^{e}_n, {B^{e}}^{T} v^{e}_n)}_{N_v}$. $V^{e}_J$ stands for the eigenvector subset indexed by $J$, i.e., $V^{e}_J \equiv \{{B^{e}}^{T} v^{e}_n\}_{n \in J}$. In addition, $e_{k,N_v}^{e}=\sum\nolimits_{{\vert J \vert =k}\wedge J \subseteq \{1,2,\ldots,N_v\}} \prod_{n \in J} \lambda_n$ is the  $k$-th elementary symmetric polynomials on eigenvalues, which is equivalent to the normalization constant in Eq. (\ref{ExpProb}) and can be computed by a recursive algorithm. %in time $O({\vert E \vert k})$ \cite{11_kulesza2011k}. 
Lastly, $P^{V^{e}_J}$ denotes an elementary DPP with marginal kernel $K^{V^{e}_J}=\sum\nolimits_{n \in V^{e}_J}  {B^{e}}^T v^{e}_n {v^{e}_n}^T {B^{e}}$.

\subsection{Implicit Relation Modelling} \label{imRM}

Building on our work on learning implicit couplings \cite{Coupling2015cao} and its application in inferring implicit rules with item dependencies \cite{wang2017inferring}, here we model and measure the implicit relations between paths in the DPP-based NSP graph $\mathcal{G}$ in terms of their indirectly linked paths. This measures the implicit relations between NSP patterns in terms of their non-occurring NSP patterns through the DPP-based NSP graph. 

\begin{definition}[Implicit Pattern Significance]
    In $\mathcal{G}$, pattern $Y_i$ (corresponding to  path $\bm{Y_i}$) \textit{is highly implicitly significant} if the items in $Y_i$ are highly dependent on other NSP itemsets $Z$ (corresponding to their paths $\bm{Z}$), indicating that the items in $Y_i$ may be relatively less likely co-occur but have a high probability of co-occurring with the items in $Z$. 
\end{definition}

Here, NSP itemsets $Z$ are \textit{link sequences} serving as a bridge to indirectly associate the items in $Y_i$, where the items are \textit{implicitly related with} those in $Z$. 

Further, we model the \textit{implicit pattern relations} between a pair of NSP patterns.
\begin{definition}[Implicit Pattern Relation]
Patterns $Y_i$ and $Y_j$ (corresponding to paths $\bm{Y_i}$ and $\bm{Y_j}$ in $\mathcal{G}$) are implicitly related if they are explicitly dependent (co-occurring) on other itemsets $Z$, represented as $Y_i \oplus Y_j \vert Z$. This indicates the items in $Y_i$ and $Y_j$ likely co-occur with those in $Z$. 
\end{definition}
If $Y_i$ and $Y_j$ share a larger number of link itemsets in the DPP-based NSP graph and have a stronger dependence on these itemsets, they are more highly implicitly related. As a result, they are less likely to co-appear in the representative NSP subset.  

Following the eNEMI measure for quantifying explicit pattern relations (in Section \ref{ExpRelMod}), the \textit{implicit relations} between an item $i$ and NSP subsequence $Z$ is measured by the \textit{implicit normalized element-wise mutual information} (iNEMI):
\begin{definition}
    The \textit{implicit normalized element-wise mutual information} (iNEMI) between $i$ and $Z$, i.e., $iNEMI(i, Z)$, measures the \textit{implicit relations} between an item $i$ and other non-co-occurring NSP itemsets $Z$. 
    \begin{small}
    \begin{equation}
        iNEMI(i,Z) =\frac{h(i)+h(Z)-h(i,Z)}{h(i,Z)} 
    \end{equation}
    \end{small}
\end{definition}
$h(i)=-log p(i)$ and $h(Z)=-log p(Z)$, $p(i)$ and $p(Z)$ are the marginal probabilities of $i$ and $Z$ in the DPP-based NSP graph, and $p(i,Z)$ is their joint probability.

\begin{definition}[Implicitly Dependent Itemsets]
    If the implicit relations between an item $i$ and an NSP itemsets $Z$ satisfy a given threshold $\epsilon \leq 0$, i.e., $iNEMI(i, Z) > \epsilon$, then all the dependent NSP itemsets of $i$ constitute a group of \textit{implicitly dependent itemsets}, denoted as $A_{i}=\{Z \vert iNEMI(i, Z) > 0\}$.
\end{definition}
$Z$ is also called the \textit{dependent itemsets} of item $i$. We also call $A_{i}$ the \textit{implicitly dependent subsequence group}.

Further, given an NSP subsequence $I$, the intersection set of all the dependent subsequence groups of the items in $I$ constitutes the \textit{link groups} of $I$, denoted as $G_{I} \equiv \cap_{i \in I} A_{i}$. Lastly, we define the \textit{implicit relation strength}.
\begin{definition}[Implicit Relation Strength]
Given each shared dependent subsequence $H \in G_{I}$, the \textit{conditional implicit relation strength} (CIRS) of $I$ over $H$ is defined as the minimum of the iNEMI between each item \textit{i} in $I$ and subsequence $H$, i.e.,
\begin{equation}
    CIRS(I \vert H) \equiv \min_{i} \{iNEMI(i,H) \vert i \in I \}
\end{equation}
and the \textit{implicit relation strength} (IRS) of $I$ is defined as the sum of its CIRS over all dependent itemsets, i.e., 
\begin{equation}
    IRS(I) \equiv \frac{\sum\nolimits_{H \in G_{I}} CIRS(I \vert H)}{\vert G_{I} \vert}
\end{equation}
\end{definition}
%$IRS(I) \equiv \sum\nolimits_{H \in G_{I}} CIRS(I \vert H)$.

Accordingly, we define the \textit{implicit pattern quality} of an NSP pattern $Y_i$.
\begin{definition}[Implicit Pattern Quality]
    In $\mathcal{G}$, the \textit{implicit pattern quality} of a path $\bm{Y_i}$ corresponding to pattern $Y_i$ is defined as the IRS of the corresponding itemsets transformed from $Y_i$, denoted as set $S_i^{q}(Y_i)$.
\end{definition}
However, on one hand, an NSP pattern may correspond to multiple implicit pattern quality terms. On the other hand, some long-size patterns may not share any dependent itemsets, i.e., $G_{q_i(\bm{y_i})} = \varnothing$. We further define the \textit{maximum implicit relation strength}.

\begin{definition}[Maximum Implicit Relation Strength]
    The \textit{maximum implicit relation strength} of a path $\bm{Y_i}$ corresponding to pattern $Y_i$ is defined as the maximum IRS of the subsets $S^{\prime}$ with the largest size, i.e., $q_i^{m}(\bm{Y_i})$, calculated as follows:
    \begin{small}
    \begin{equation}
        q_i^{m}(\bm{Y_i})= max\{IRS(S^{\prime}) \vert \{ \arg \max_{size(S^{\prime})} \vert {S^{\prime} \in \mathcal{S}} \wedge{S^{\prime} \subseteq S_i^{q}(\bm{Y_i})}  \} \}
    %q_i^{m}(\bm{y_i})= max\{IRS(S^{\prime}) \vert \{ \arg \max_{size(S^{\prime})} \vert {S^{\prime} \in \mathcal{S}} \wedge {S^{\prime} \subseteq flat(\bm{y_i})}  \} \}
    \end{equation}
    \end{small}
\end{definition}
$\mathcal{S}$ is the collection of the \textit{implicitly related itemsets} (IRI), which are the itemsets with high IRS and can be discovered by an adapted IRRMiner algorithm \cite{wang2017inferring}. The itemset $\{ argmax_{size(S^{\prime})} \vert {S^{\prime} \in \mathcal{S}} \wedge {S^{\prime} \subseteq S_i^{q}(\bm{Y_i})}\}$ contains the subsets of the IRI, in which each itemset is a subset of $S_i^{q}(\bm{Y_i})$ with the largest size. $q_i^{m}(\bm{Y_i})$ is the highest IRS of these itemsets, which thus measures the implicit quality of the major items in $\bm{Y_i}$ w.r.t. their related link itemsets.

The implicit diversity feature vector of pattern $Y_i$ is its normalized feature vector $\phi_i(\bm{Y_i})=Norm(d^{i}(\bm{Y_i}))$ corresponding to the path $\bm{Y_i}$ in the graph $\mathcal{G}$. $d^{i}(\bm{Y_i}) \in R^{\vert H_{S} \vert}$ and $H_{S}$ is the set of all dependent itemsets. $d^{i}(\bm{Y_i})$ quantifies the implicit dependency of pattern $Y_i$ w.r.t. all potential \textit{link itemsets}. $\phi_i(\bm{Y_i})^{T}\phi_i(\bm{Y_j}) \in [-1,1]$ measures the implicit similarity between $\bm{Y_i}$ and $\bm{Y_j}$ in the DPP graph. 

In this paper, the $k$-th component $d^{i}_{k}(\bm{Y_i}) \in {d}^{i}(\bm{Y_i})$ is built as the conditional IRS of $S_i^{q}(\bm{Y_i})$ on dependent itemset $H_k \in H_{S}$, i.e., $d^{i}_{k}(\bm{Y_i}) \equiv CIRS(S_i^{q}(\bm{Y_i}) \vert H_k)$. Accordingly, the production between the implicit diversity vectors of two patterns is proportional to the fraction of dependent itemsets they share. %Hence, $d^{i}(\bm{y_i})$ quantifies the explicit dependency of pattern $\bm{y_i}$ w.r.t. all potential \textit{link itemsets}, and $\phi_i(\bm{y_i})^{T}\phi_i(\bm{y_j}) \in [-1,1]$ works as the implicit similarity between patterns $\bm{y_i}$ and $\bm{y_j}$. 
%In addition, the implicit diversity feature vector of pattern $\bm{y_i}$ is constructed as $\phi_i(\bm{y_i})=Normalization(d^{i}(\bm{y_i}))$, where $d^{i}(\bm{y_i}) \in R^{\vert H_{S} \vert}$ and $H_{S}$ is the set of all dependent itemsets. Here, the k-th component $d^{i}_{k}(\bm{y_i}) \in {d}^{i}(\bm{y_i})$ is defined as the conditional IRS of $flat(\bm{y_i})$ on dependent itemset $H_k \in H_{S}$, i.e., $d^{i}_{k}(\bm{y_i}) \equiv CIRS(flat(\bm{y_i}) \vert H_k)$. Hence, $d^{i}(\bm{y_i})$ quantifies the explicit dependency of pattern $\bm{y_i}$ w.r.t. all potential \textit{link itemsets}, and $\phi_i(\bm{y_i})^{T}\phi_i(\bm{y_j}) \in [-1,1]$ works as the implicit similarity between patterns $\bm{y_i}$ and $\bm{y_j}$.
%Here $H$ is the set of all potential link itemset. %In addition, the implicit diversity feature vector of pattern $\bm{y_i}$ is constructed as $\phi_i(\bm{y_i})=Normalization(\vec{\hat{d}}(\bm{y_i}))$, where $\vec{\hat{d}}(\bm{y_i})  \in R^{\vert H \vert}$ and $\hat{d}_k(\bm{y_i}) \in \vec{\hat{d}}(\bm{y_i})$ is defined as the conditional IRS of $flat(\bm{y_i})$ and $H_k \in H$, i.e., $CIRS(flat(\bm{y_i}) \vert H_k)$. Here $H$ is the set of all potential link itemset.

Accordingly, the implicit probability of an NSP subset $Y \subseteq \mathcal{Y}$ can be computed by a \textit{k}-DPP-based model, which is as follows:
%\begin{small}
\begin{equation} \label{ImpProb}
    P^k_i(Y) = \frac{det(L^{i}_{Y})}{\sum\nolimits_{\vert Y^{\prime} \vert=k} det(L^{i}_{Y^{\prime}})}
\end{equation}
%\end{small}

Similar to the conversion in Section \ref{ExpRelMod}, kernel $L^{i}$ can be rewritten as $L^{i}_{ij}={B^{i}}^{T} B^{i}$, where the columns of  $B^{i}$ are given by $B^{i}_{k}=q_i(\bm{Y_k})\phi_i(\bm{Y_k})$, and the dual representation of $L^{i}$ is constructed as $C^{i} = {B^{i}}{{B^{i}}^T}$. Assume that $(\lambda^{i}_n, v^{i}_n)$ is an eigenvalue/eigenvector pair of $C^{i}$, then $(\lambda^{i}_n, {B^{i}}^{T} v^{i}_n)$ is the corresponding pair of $L^{i}$. Then, Eq. (\ref{ImpProb}) is rewritten as follows:

\begin{small}
\begin{equation} \label{ImpProbFor} 
    P^k_i(Y)= \frac{1}{e_{k,{N_v}}^{i}} \sum\limits_{{\vert J \vert =k} \wedge J \subseteq \{1,2,\ldots,N_v\}} P^{V^{i}_J}(Y) \prod\limits_{n \in J} \lambda^{i}_n
\end{equation}
\end{small}
Here, $P^{V^{i}_J}(Y)$ stands for an elementary DPP with marginal kernel $K^{V^{i}_J}=\sum\nolimits_{n \in V^{i}_J}  {B^{i}}^T v^{i}_n {v^{i}_n}^T {B^{i}}$.

\subsection{Overall Relations-based NSP Selection}
%Similar to the application of DPP in other areas \cite{4_yuan2016discovering}, 
To identify the representative NSP subsets, we measure the overall probability of a subset $Y$ by integrating the above explicit and implicit relations-oriented subset probabilities as follows:

\begin{equation} \label{OverallPro}
    P^k(Y) = w_{e} \times P^k_e(Y) + w_{i} \times P^k_i(Y)
    %P^k(Y) = \frac{\bar{freq}(Y) \times P^k_e(Y) +\bar{IRS}(flat(Y)) \times P^k_i(Y)}{\bar{freq}(Y) + \bar{IRS}(flat(Y))}
\end{equation}
$w_{e}$ and $w_{i}$ are the parameters to balance between $P^k_e(Y)$ and $P^k_i(Y)$, and $w_{e}+w_{i}=1$. Here, we adopt  $w_{e} =\frac{\bar{freq}(Y)}{\bar{freq}(Y) + \bar{IRS}(q_i(Y)} $ and $w_{i} =\frac{\bar{IRS}(q_i(Y))}{\bar{freq}(Y) + \bar{IRS}(q_i(Y)}$. $\bar{freq}(Y)$ is the average frequency of the patterns in $Y$,  $\bar{IRS}(q_i(Y))$ is the average IRS of the flat itemsets transformed from the patterns in $Y$. In this way, the overall probability $P^k(Y)$ is more affected by the major relations of patterns in subset $Y$. Substituting Eqs. (\ref{ExpProFor}) and (\ref{ImpProbFor}) in Eq. (\ref{OverallPro}), we obtain the overall probability of a subset $Y$.

\begin{small}
\begin{equation} \label{OverallProbFor} 
    P^k(Y)= \sum\limits_{d \in \{e,i\}} \sum\limits_{{\vert J \vert =k} \wedge J \subseteq \{1,2,\ldots,N_v\}} \frac{w_{e} P^{V^{d}_J}(Y)}{e_{k,{N_v}}^{d}}   \prod\limits_{n \in J} \lambda^{d}_n
\end{equation}
\end{small}
$P^k(Y)$ will be used to select the $k$-size subset in the following section.

\subsection{The EINSP Algorithm}
EINSP implements the DPP-based actionable NSP discovery process in Fig. \ref{framework}. EINSP selects a representative NSP subset based on its overall subset probability $P^k(Y)$. Per Eq. \ref{OverallProbFor}, $P^k(Y)$ is modeled as a mixture of element-wise DPPs (similar to the subset selection proved in \cite{0_kulesza2012determinantal}). EINSP samples the subset within two main loops corresponding to two phases of sampling in Algorithm \ref{code}. In the first loop, a subset of $k$ eigenvectors is selected, where the probability of selecting each eigenvector depends on its associated eigenvalue. A $k$-size index subset $J$ is sampled by $P(J)=\sum\nolimits_{d \in \{e,i\}} \frac{w_{d}}{e_{k,N_v}^{d}} \prod\nolimits_{n \in J} \lambda^{d}_n$. The marginal probability of index $n \in J$ is as follows:
\begin{small}
\begin{equation} \label{NProb}
    P(n \in J)=\sum\nolimits_{d \in \{e,i\}} w_{d}\lambda^{d}_{n}\frac{e_{k-1,n-1}^{d}}{e_{k,n}^{d}}
\end{equation}
\end{small}

With the dual representation,  sets $V^{e}$ and $V^{i}$ of the eigenvectors of kernels $L^{e}$ and $L^{i}$ are represented by their corresponding sets of eigenvectors of $C^{e}$ and $C^{i}$. The eigenvectors are denoted as $\hat{V}^{e}$ and $\hat{V}^{i}$, with the mapping $V^{e}=\{{B^e}^{T}\hat{v}^e \vert \hat{v}^e \in \hat{V}^{e} \}$ and $V^{i}=\{{B^i}^{T}\hat{v}^i \vert \hat{v}^i \in \hat{V}^{i} \}$. Consequently, for any two eigenvectors $\hat{v}_{i}^e, \hat{v}_{j}^e \in \hat{V}^{e}$, we have ${\hat{v_{i}^e}}^{T} {\hat{v}_{j}^e}={\hat{v_{i}^e}}^{T} C^{e} {\hat{v}_{j}^e}$. Accordingly, the normalization of the vectors in $V^{e}$ and $V^{i}$ can be computed  using only their preimages in $\hat{V}^{e}$ and $\hat{V}^{i}$ \cite{0_kulesza2012determinantal} by updating $\hat{v}^e_n \gets \{\frac{\hat{v}^e_n}{{\hat{v^e_n}}^{T}C^{e}\hat{v}^e_n}\}$ and $\hat{v}^i_n \gets \{\frac{\hat{v}^i_n}{{\hat{v^i_n}}^{T}C^{i}\hat{v}^i_n}\}$. 

In the second phase, a subset $Y$ is generated from the selected eigenvectors. On each iteration of this second loop, the cardinality of $Y$ increases by one and the dimensionality of $\hat{V}^{e}$ and $\hat{V}^{i}$ is reduced by one. Here, $e_j$ is the $j$-th standard basis vector, which is all zeros except for a one in the $j$-th position. During each iteration, EINSP selects a pattern $\bm{Y_{j}}$ according to the following distribution:

\begin{small}
\begin{equation} 
\begin{split} \label{code_select} 
    P(\bm{Y_{j}})&=w_{e}\frac{1}{\vert V^e \vert}\sum\limits_{\hat{v}^e \in \hat{V}^e} ({v^e}^{T}e_j)^2 + w_{i}\frac{1}{\vert V^i \vert}\sum\limits_{\hat{v}^i \in \hat{V}^i} ({v^i}^{T}e_j)^2 \\
    &=\sum\limits_{d \in \{e,i\}} \sum\limits_{\hat{v}^d \in \hat{V}^d} \frac{w_{d}}{\vert V^d \vert} q_{d}^2(\bm{Y_j})({\hat{v_{j}^d}}^{T} \phi_d(\bm{Y_j}))^2
    % + w_{i}\frac{1}{\vert V^i \vert}\sum\limits_{\hat{v}^i \in \hat{V}^i} q_{i}^2(\bm{y_j})({\hat{v_{j}^i}}^{T} \phi_i(\bm{y_j}))^2 
    %&=w_{e}\frac{1}{\vert V^e \vert}\sum\nolimits_{\hat{v}^e \in \hat{V}^e} (({B^e}^{T}\hat{v}^e)^{T}e_j)^2 + w_{i}\frac{1}{\vert V^i \vert}\sum\nolimits_{\hat{v}^i \in \hat{V}^i} (({B^i}^{T}\hat{v}^i)^{T}e_j)^2
\end{split}
\end{equation}
\end{small}

Algorithm \ref{code} summarizes the working mechanism and process of the proposed EINSP method for actionable NSP selection. % pseudo-code 

{\small
\begin{algorithm}[ht]%[h]  
\caption{The Pseudo-code of EINSP for Actionable NSP Selection} 
\renewcommand{\algorithmicrequire}{\textbf{Input:}}
\renewcommand{\algorithmicensure}{\textbf{Output:}}
\label{code}
\begin{algorithmic}[1]
\Require  NSP Collection $\mathcal{Y}=\{\bm{Y_1},\bm{Y_2},\ldots,\bm{Y_N}\}$, cardinality $k$ %eigenvalue/eigenvector pairs $\{(\lambda^{e}_n, {\hat{v}}^{e}_n)\}_{n=1}^{N_v}$, $\{(\lambda^{i}_n, {\hat{v}}^{i}_n)\}_{n=1}^{N_v}$ of dual expression $C^e$ and $C^i$ respectively, size $k$ % eigenvalue/eigenvector pairs ${(\lambda_k, v_k)}_{N_v}$ of dual expression $C$
\Ensure Representative Subset $Y$ 
\State  Map $\mathcal{Y}$ to a directed graph $\mathcal{G}$ per the NSP graph construction
\State  Construct dual representations $C^e$ and $C^i$ and compute their eigenvalue/eigenvector pairs $\{(\lambda^{e}_n, {\hat{v}}^{e}_n)\}_{n=1}^{N_v}$ and $\{(\lambda^{i}_n, {\hat{v}}^{i}_n)\}_{n=1}^{N_v}$ of $C^e$ and $C^i$, respectively
\State  $J \gets \varnothing$ 
\For{$n = 1,2,\ldots,N_v$} %\For{$n = 1,2,\ldots,min\{{N_e},{N_i}\}$}
\If{$u \sim U[0,1] < \sum\nolimits_{d \in \{e,i\}} w_{d}\lambda^{d}_{n}\frac{e_{k-1,n-1}^{d}}{e_{k,n}^{d}}$} \State  $J \gets J \cup \{n\}$ \State  $k \gets k-1$ 
\If {$k=0$}
    \textbf{break}  
\EndIf
\EndIf
\EndFor
\State  $\hat{V}^e \gets \{\frac{\hat{v}^e_n}{{{\hat{v^e_n}}^{T}} C^{e} \hat{v}^e_n}\}_{n \in J}$
\State  $\hat{V}^i \gets \{\frac{\hat{v}^i_n}{{\hat{v^i_n}}^{T}C^{i}\hat{v}^i_n}\}_{n \in J}$
\State  $Y \gets \varnothing$
\While{$V \not= \varnothing$}
\State Select $\bm{Y_{j}}$ from $\mathcal{Y}$ with $P(\bm{Y_{j}})=\sum\nolimits_{d \in \{e,i\}} \sum\nolimits_{\hat{v}^d \in \hat{V}^d} \frac{w_{d}}{\vert V^d \vert} q_{d}^2(\bm{Y_j})({\hat{v_{j}^d}}^{T} \phi_d(\bm{Y_j}))^2$ 
\State  $Y \gets Y \cup \bm{Y_{j}}$
\State  $V_e \gets V_{e,\perp}$, where $\{{B^{e}}^{T}v_{e} \vert v_{e} \in V_{e,\perp}\}$ is an orthonormal basis for the subspace of $V_e$ orthogonal to $e_j$
\State  $V_i \gets V_{i,\perp}$, where $\{{B^{i}}^{T}v_{i} \vert v_{i} \in V_{i,\perp}\}$ is an orthonormal basis for the subspace of $V_i$ orthogonal to $e_j$
\EndWhile
\State \Return{$Y$}
\end{algorithmic}
\end{algorithm}
}

\section{Complexity Analysis and Comparison}	
\label{sec:compana}

Here, we analyze the computational complexity of EINSP and the baseline methods in Section \ref{secEA} for comparison. The EINSP computational complexity is determined by its four constituent parts. First, the NSP graph construction in Section \ref{GraphCon} maps an NSP pattern set $\mathcal{Y}$ to a DPP graph $\mathcal{G}$; its computational complexity is mainly sensitive to the size of all elements $E$ in the NSP set $\mathcal{Y}$ and the pattern number in the set $|\mathcal{Y}|$, i.e., $O(|E||\mathcal{Y}|)$. Second, the explicit relation modeling in Section \ref{ExpRelMod} generates a DPP-based graph representation $C^{e}$ with the time complexity $O({\vert E \vert}^2{\vert \mathcal{Y} \vert})$; it then generates a subset $Y$ forming the k-size SDPP (i.e., k-SDPP) and calculates their probability in Eq. (\ref{ExpProFor}) by a recursive algorithm \cite{11_kulesza2011k} with time $O({\vert E \vert k})$. Third, the implicit relation modeling in Section \ref{imRM} measures the implicit pattern relations  sensitive to the item number $|I|$ and itemset (element) number $|E|$ with complexity $O(|I||E|)$; it then calculates the implicit probability of subset $Y$ in Eq. (\ref{ImpProbFor}) sensitive to items $I$ and $k$-size subsets with complexity $O({\vert I \vert k})$ (similar to the calculation of explicit probability). Lastly, the overall relation-based NSP selection combines the explicit and implicit probability in the second and third steps for NSP subset selection. Consequently, the overall computational complexity of EINSP can be estimated as $O(|E||\mathcal{Y}| + {\vert E \vert}^2{\vert \mathcal{Y} \vert} + {\vert k \vert E} + |I||E| + { \vert k \vert I})$.   

Further, we briefly explain the computational complexity of the baseline methods. First, top-K selection selects top-k frequentest patterns from the pattern set $\mathcal{Y}$ with complexity $O(|\mathcal{Y}|)$. Second, SAPNSP applies e-NSP with complexity $O(e-NSP)$ to generate the NSP set with pattern number $|e-NSP|$ and then select its subset with complexity $O(O(e-NSP) + |e-NSP|)$ (Interested readers can refer to \cite{e-NSP2} for $O(e-NSP)$ and $|e-NSP|$). Third, the k-means method applies k-means clustering to select diverse patterns with complexity $O(|\mathcal{Y}|^2)$ over the NSP number $|\mathcal{Y}|$. Lastly, the EINSP variant k-SDPP only involves explicit relations with complexity $O(|E||\mathcal{Y}| + {\vert E \vert}^2{\vert \mathcal{Y} \vert} + {\vert k \vert E})$. 

The above estimation shows these strategies serve different purposes and are highly incomparable. Their divided computational costs do not reflect their capacity. Empirically comparing their computational costs in experiments does not give much insight about how and why they work, which could instead generate misleading indication. Therefore, in the following evaluation, we focus on evaluating the capacity of the EINSP design in selecting more actionable patterns. EISNP and its baseline strategies are evaluated in terms of their capacity of (1) pattern (element and item) coverage and diversity (Section \ref{PatCov}); (2) pattern size/length and diversity clustering effect (Section \ref{PatSize}); (3) learning implicit pattern relations (Section \ref{PatRel}); (4) handling data with complex data factor combinations (Section \ref{MetSen}); (5) ablation study (in all sections \ref{PatCov} to \ref{MetSen} on k-SDPP with EINSP); (6) learning 17 synthetic data with different complexities (see Tables \ref{MS1} and \ref{MS2}); and (7) learning six highly divided real-life data (Figs. \ref{SeqCovFig}-\ref{PatRelFig}). These comprehensive evaluations complement the above complexity analysis with a deep insight about the soundness, robustness, scalability and flexibility of our approach.

\section{Experiments and Evaluation}	
\label{secEA}

The empirical analysis of the proposed EINSP  in comparison with four baselines is undertaken on six real-life datasets and 17 synthetic datasets. Here, we first introduce the datasets and baseline methods, and then evaluate the EINSP performance in terms of multiple evaluation perspectives.

\subsection{Experiment Setup}

\subsubsection{Datasets} \label{datasets}
We adopt the following six real-life datasets to evaluate the efficiency of EINSP  against the baselines.

\begin{itemize}
    \item Dataset 1 (DS1): A UCI dataset with 989,818 anonymously ordered webpage visits to MSNBC.com \cite{e-NSP2}. %Visits were recorded at the page category and in a temporal order \cite{e-NSP2}.
    \item Dataset 2 (DS2): A real-life health insurance claim sequential dataset \cite{e-NSP2} with an average of 21 elements per sequence and two items per element, 5,269 data sequences, and 340 divergent items. %The file size is around 5M.
    \item Dataset 3 (DS3): A chain-store real-life dataset containing 46,086 distinct items and 1,112,949 transactions \cite{NU-MineBench}.% which is widely adopted for testing sequential pattern, especially for utility-based pattern.
     \item Dataset 4 (DS4): A KDD-CUP 2000 dataset\footnote{\url{http://www.philippe-fournier-viger.com/spmf/datasets/BMS1_spmf}} with 59,601 e-commerce clickstream sequences, 497 distinct items, and an average of 2.42 items per sequence with a standard deviation of 3.22. %In this dataset, there are some long sequences.
     \item Dataset 5 (DS5): Another KDD-CUP 2000 dataset\footnote{\url{http://www.philippe-fournier-viger.com/spmf/datasets/BMS2.txt}} with 77,512 click-stream sequences and 3,340 distinct items, an average of 4.62 items per sequence with a standard deviation of 6.07 items.
     %\item Dataset 4 (DS4), is a dataset of 20,450 sequences of click stream data from the website of FIFA World Cup 98. It has 2,990 distinct items (webpages). The average sequence length is 34.74 items with a standard deviation of 24.08 items. This dataset was created by processing a part of the web logs from the World Cup web log.
     \item Dataset 6 (DS6): A FIFA World Cup'98 dataset\footnote{\url{http://www.philippe-fournier-viger.com/spmf/datasets/FIFA.txt}} with 20,450 clickstream sequences, 2,990 distinct items, and an average of 34.74 items per sequence with a standard deviation of 24.08 items. %This dataset was created by processing a part of the web logs from the World Cup web log  from the world cup \footnote{\url{http://ita.ee.lbl.gov/html/contrib/WorldCup.html}}.
     %is generated to evaluate the scalability of the algorithms on multiple datasets with different data factors, which is then expanded to 11 synthetic datasets, named DS10.X (X=0,\ldots,10), by adjusting one factor once.
     %\item Dataset 6 (DS6), is a dataset of 20,450 sequences of click stream data from the website of FIFA World Cup 98 \footnote{\url{http://www.philippe-fournier-viger.com/spmf/datasets/FIFA.txt}}. It has 2,990 distinct items (webpages). The average sequence length is 34.74 items with a standard deviation of 24.08 items.
\end{itemize}

We also generate another 17 synthetic datasets using the IBM data generator \cite{AprioriAll} to evaluate the sensitivity of EINSP and the baselines on different data factors, i.e., how the different characteristics of data influence the effectiveness of each method, following the data factor analysis in \cite{e-NSP2}. Table \ref{SeqDsDetail} describes these datasets in terms of six data factors. The base dataset is $C10\_T6\_S8\_I8\_DB10k\_N0.1k$, which is further expanded to others by adjusting one factor (in boldface) once.  %The adopted synthetic datasets are extended from the base dataset by tuning one factor and the differentiator are marked by bolding the distinct factor.
Here, \textit{C} is the average number of elements per sequence, \textit{T} is the average number of items per element, \textit{S} is the average length of potentially maximal frequent positive sequences; \textit{I} is the average number of items per element in potentially maximal frequent positive sequences; 
\textit{DB} is the number of data sequences in a sequence dataset, and \textit{N} is the number of distinct items. 

{\small
\begin{table*}%[!t]
%\normalsize
\renewcommand{\arraystretch}{1.3}
\caption{Synthetic Sequence Datasets w.r.t. Different Data Factors}
\label{SeqDsDetail}
\centering
\scalebox{0.8}[0.8]{
%\begin{tabular}{c | c | c}
\begin{tabular}{|p{55pt} | p{110pt} | p{190pt}|}
%\begin{tabular}{c|c|c|c|c|c}
\hline
\bfseries Data Factor & \bfseries Dataset Name & \bfseries  Data Factor Adjustment \\
\hline \hline
Base Dataset & C10\_T6\_S8\_I8\_DB10k\_N0.1k & $C=10$, $T=6$, $S=8$, $I=8$, $DB=10k$, $N=0.1k$ \\ \hline
C=6 & \textbf{C6}\_T6\_S8\_I8\_DB10k\_N0.1k & $C=6$, $T=6$, $S=8$, $I=8$, $DB=10k$, $N=0.1k$ \\ \cline{2-3}
C=8 & \textbf{C8}\_T6\_S8\_I8\_DB10k\_N0.1k & $C=8$, $T=6$, $S=8$, $I=8$, $DB=10k$, $N=0.1k$ \\ \cline{2-3}
C=12 & \textbf{C12}\_T6\_S8\_I8\_DB10k\_N0.1k & $C=12$, $T=6$, $S=8$, $I=8$, $DB=10k$, $N=0.1k$ \\ \cline{2-3}
C=14 & \textbf{C14}\_T6\_S8\_I8\_DB10k\_N0.1k & $C=14$, $T=6$, $S=8$, $I=8$, $DB=10k$, $N=0.1k$ \\ \hline
T=4 & C10\_\textbf{T4}\_S8\_I8\_DB10k\_N0.1k & $C=10$, $T=4$, $S=8$, $I=8$, $DB=10k$, $N=0.1k$ \\ \cline{2-3}
T=8 & C10\_\textbf{T8}\_S8\_I8\_DB10k\_N0.1k & $C=10$, $T=8$, $S=8$, $I=8$, $DB=10k$, $N=0.1k$ \\ \cline{2-3}
T=10 & C10\_\textbf{T10}\_S8\_I8\_DB10k\_N0.1k & $C=10$, $T=10$, $S=8$, $I=8$, $DB=10k$, $N=0.1k$ \\ \cline{2-3}
T=12 & C10\_\textbf{T12}\_S8\_I8\_DB10k\_N0.1k & $C=10$, $T=12$, $S=8$, $I=8$, $DB=10k$, $N=0.1k$ \\ \hline
%S\_1 & C10\_T6\_S4\_I8\_DB10k\_N0.1k & $C=10$, $T=6$, $S=4$, $I=8$, $DB=10k$, $N=0.1k$ \\ \cline{2-3}
%S\_2 & C10\_T6\_S6\_I8\_DB10k\_N0.1k & $C=10$, $T=6$, $S=6$, $I=8$, $DB=10k$, $N=0.1k$ \\ \cline{2-3}
%S\_3 & C10\_T6\_S10\_I8\_DB10k\_N0.1k & $C=10$, $T=6$, $S=10$, $I=8$, $DB=10k$, $N=0.1k$ \\ \cline{2-3}
%S\_4 & C10\_T6\_S12\_I8\_DB10k\_N0.1k & $C=10$, $T=6$, $S=12$, $I=8$, $DB=10k$, $N=0.1k$ \\ \hline
%I\_1 & C10\_T6\_S8\_I4\_DB10k\_N0.1k & $C=10$, $T=6$, $S=8$, $I=4$, $DB=10k$, $N=0.1k$ \\ \cline{2-3}
%I\_2 & C10\_T6\_S8\_I6\_DB10k\_N0.1k & $C=10$, $T=6$, $S=8$, $I=6$, $DB=10k$, $N=0.1k$ \\ \cline{2-3}
%I\_3 & C10\_T6\_S8\_I10\_DB10k\_N0.1k & $C=10$, $T=6$, $S=8$, $I=10$, $DB=10k$, $N=0.1k$ \\ \cline{2-3}
%I\_4 & C10\_T6\_S8\_I12\_DB10k\_N0.1k & $C=10$, $T=6$, $S=8$, $I=12$, $DB=10k$, $N=0.1k$ \\ \hline
DB=20k & C10\_T6\_S8\_I8\_\textbf{DB20k}\_N0.1k & $C=10$, $T=6$, $S=8$, $I=8$, $DB=20k$, $N=0.1k$ \\ \cline{2-3}
DB=30k & C10\_T6\_S8\_I8\_\textbf{DB30k}\_N0.1k & $C=10$, $T=6$, $S=8$, $I=8$, $DB=30k$, $N=0.1k$ \\ \cline{2-3}
DB=40k & C10\_T6\_S8\_I8\_\textbf{DB40k}\_N0.1k & $C=10$, $T=6$, $S=8$, $I=8$, $DB=40k$, $N=0.1k$ \\ \cline{2-3}
DB=50k & C10\_T6\_S8\_I8\_\textbf{DB50k}\_N0.1k & $C=10$, $T=6$, $S=8$, $I=8$, $DB=50k$, $N=0.1k$ \\ \hline
N=0.2k & C10\_T6\_S8\_I8\_DB10k\_\textbf{N0.2k} & $C=10$, $T=6$, $S=8$, $I=8$, $DB=10k$, $N=0.2k$ \\ \cline{2-3}
N=0.3k & C10\_T6\_S8\_I8\_DB10k\_\textbf{N0.3k} & $C=10$, $T=6$, $S=8$, $I=8$, $DB=10k$, $N=0.3k$ \\ \cline{2-3}
N=0.4k & C10\_T6\_S8\_I8\_DB10k\_\textbf{N0.4k} & $C=10$, $T=6$, $S=8$, $I=8$, $DB=10k$, $N=0.4k$ \\ \cline{2-3}
N=0.5k & C10\_T6\_S8\_I8\_DB10k\_\textbf{N0.5k} & $C=10$, $T=6$, $S=8$, $I=8$, $DB=10k$, $N=0.5k$ \\ \hline
\end{tabular}
}
\end{table*}
}

\subsubsection{Baseline Methods}
To the best of our knowledge, none of the existing work is available to discover the actionable NSP subset by both satisfying pattern significance (quality) and diversity and involving explicit and implicit pattern relations as in our work. We specify the baseline methods for the following test: (1) pattern quality based on frequentist by Top-k NSP selection and SAPNSP; (2) subset diversity by k-means; and (3) EINSP ablation study by k-SDPP. 
%(1) a Top-k NSP selection method and SAPNSP that select NSP subsets only considering pattern quality (support or contribution); (2) k-means method that discovers NSP subsets only in terms of subset diversity; (3) a DPP-based k-SDPP that discovers the subset only in terms of the explicit relations between NSP itemsets, k-SDPP is an EINSP variant for ablation study.  
%So far SAPNSP is the only method proposed to discover an optimal subset from a huge NSP collection \cite{SAPNSP}. 
%. Different from the mechanism of our proposed method that evaluate each sequence in terms of both its importance and its diversity information, SAPNSP proposes 
%which only evaluates the importance of each single pattern in terms of its support and intra-sequence correlation. 

\begin{itemize}
    \item Top-k selection (Top-k for short): A simple baseline to select the top-k patterns with the highest frequency, which evaluates each pattern only in terms of its \textit{support}.
    \item SAPNSP \cite{DongLXW15}: This is the only method available to select a subset of top-k patterns from an NSP collection. It applies the \textit{contribution} measure proposed in \cite{zhao2009mining} to evaluate the importance of each pattern in terms of its frequency and intra-sequence overlap between its prefix and last element.
    \item k-means baseline (k-means): A diversity-oriented baseline to apply k-means clustering for NSP collection by using the proposed explicit diversity (Eq. (\ref{eq:eNEMI})) to measure the distance between a pair of patterns and then select the most frequent patterns in each cluster to form a k-size subset\footnote{Interested readers may also refer to \cite{3_gillenwater2012discovering} for k-means-based selection.}.
    \item k-SDPP baseline (k-SDPP): A variant of EINSP, which selects a k-size representative subset by only modelling the co-occurrence-based explicit NSP relations in the DPP-based NSP graph.
    \item EINSP: The full model to select a k-size representative subset by jointly modelling both explicit and implicit NSP relations in the DPP-based NSP graph.
\end{itemize}

All the algorithms are implemented in Python, and experiments are conducted on a cluster running node with Intel Xeon W3690 (6 Core) CPU of 3.47GHz, 12GB memory and Red Hat Enterprise Linux 6.7 (64bit) OS. Sequences are constructed in each dataset per the time or order-related information in the data. These baseline methods are evaluated on the aforementioned real-life and synthetic datasets. To verify the effectiveness of EINSP in discovering a significant and diverse NSP subset, we conduct empirical comparisons between EINSP and the baselines from the perspectives of \textit{pattern coverage}, \textit{average pattern size}, and \textit{average implicit relation strength} of the selected subset in Sections \ref{PatCov}, \ref{PatSize} and \ref{PatRel}, respectively. In addition, a sensitivity comparison is conducted in Section \ref{MetSen} to test the stability of EINSP against the baselines on datasets in terms of different data factors following \cite{e-NSP2}. 

Negative-GSP \cite{Neg-GSP} is adopted to discover the complete NSP set since it is shown to be the most efficient of discovering complete NSPs. Other NSP methods including e-NSP and its specific extensions and other frequentist-based NSP methods are not selected because they work on specific settings and constraints, leading to non-complete NSPs. In fact, we are not concerned with the NSP miners as long as they can produce the complete NSP set as the input. Negative-GSP contains one parameter $min\_sup$ as a frequency constraint on guaranteeing the frequent co-occurrences of items in patterns. In Sections \ref{PatCov}, \ref{PatSize} and \ref{PatRel}, we tailor $min\_sup$ as 10\%, 20\%, 1.5\%,1.5\%, 1.5\% and 20\% empirically for the six real-life datasets and 30\% for all synthetic datasets by jointly and empirically balancing their data characteristics, resultant pattern distributions, and result comparison.

\subsection{Pattern Coverage} \label{PatCov}

The \textit{sequence coverage} and \textit{item coverage} \cite{e-NSP2,WangC19} of selected subsets measure the diversity of selected NSPs and distinct NSP items. We report the results of EINSP and the baselines in Sections \ref{SeqCov} and \ref{ItemCov}. We also compare the \textit{average item frequency} of selected NSP subsets in Section \ref{AcifCov}. Here, we first quantify the measures \textit{sequence coverage}, \textit{item coverage}, and \textit{average item frequency} for this work.

\textit{Sequence coverage} measures the diversity of the selected NSP itemsets in the entire sequence database.  
\begin{definition}[Sequence Coverage]
    The \textit{sequence coverage} of subset $S$ in dataset $D$ is denoted as $SC(S \vert D)$, which is the ratio of the data sequences in $D$ that cover at least one pattern in $S$ with respect to the size of the dataset. 
    \begin{equation}
        SC(S \vert D) \equiv \frac{\vert \{ s_{d} \vert s_{d} \in D \wedge (\exists s_{p} \in S \ s.t. \ s_{p} \subseteq s_{d}) \} \vert}{\vert D \vert}
    \end{equation}
    where $s_{d}$ is a data sequence from dataset $D$, and $s_{p}$ is a pattern from the selected subset $S$. 
\end{definition}

The sequence coverage $SC(S \vert D)$ is always much lower than the sum of the frequency of the patterns in $S$, i.e., $SC(S \vert D) \ll \sum \nolimits_{s_{p} \in S} sup(s_{p} \vert D)$. $sup(s_{p} \vert D)$ is the support of pattern $s_{p}$ in dataset $D$. This is because the cover sets of different patterns may not be disjoint, i.e., $cov(s_{p} \vert D) \bigcap cov(s_{p}^{\prime} \vert D) \not= \varnothing$ where $s_{p},s_{p}^{\prime} \in S$ and $cov(s_{p} \vert D) \equiv \{ s_{d} \vert s_{d} \in D \wedge s_{p} \subseteq s_{d}\}$.  %  $SC(S \vert D) \leqslant \sum \nolimits_{s_{p} \in S} sup(s_{p} \vert D)$

\textit{Item coverage} measures the diversity of items in the selected NSP set in the entire item set.
\begin{definition}[Item Coverage]
    The \textit{item coverage} of subset $S$ in dataset $D$ measures the ratio of the items in $D$ covered by at least one pattern in $S$ with respect to the whole item population, i.e., 
    \begin{equation}
        IC(S \vert D) \equiv \frac{\vert \{ i \vert \exists s_{d} \in D, \ s_{p} \in S \ s.t. \ i \subseteq s_{d} \wedge i \subseteq s_{p} \} \vert}{\vert I_{D} \vert}
    \end{equation}
     where $i$ is an item and $I_{D} \equiv \{ i \vert \exists s_{d} \in D \ s.t. \ i \subseteq s_{d} \}$ is the set of items covered by dataset $D$.
\end{definition}

The \textit{average item frequency} measures the frequentist significance of items covered by the selected NSP set. 
\begin{definition}[Average Item Frequency]
    The \textit{item frequency} of an item $i$ in subset $S$ is the ratio of its occurrence times in $S$ with respect to the size of $S$. Accordingly, the \textit{average item frequency} in subset $S$ is defined as
    \begin{equation}
        AF(S \vert D) \equiv \frac{1}{\vert I_{S} \vert} \sum \nolimits_{i \in I_{S}} \frac{\vert \{ s_{p} \vert s_{p} \in S \wedge i \subseteq s_{p} \} \vert}{\vert S \vert }
    \end{equation}
\end{definition}

In general, an ideal representative subset has a high \textit{sequence coverage} such that a large proportion of data sequences in the original dataset are covered by a small-scale subset. A subset with a higher \textit{item coverage} indicates more informative and diverse items covered in the subset selected from the entire itemset. A subset with a lower \textit{average item frequency} tends to be more diverse and balanced. These measures quantify the various levels of diversity of an NSP miner.

%In addition, the \textit{item coverage} of a subset $S$ in a dataset $D$ is defined as the ratio of the items in $D$ covered by the patterns in $S$ with respect to the whole item population, i.e., $IC(S \vert D) \equiv \frac{\vert \{ i \vert \exists s_{d} \in D, s_{p} \in S s.t. i \subseteq s_{d} \wedge i \subseteq s_{p} \} \vert}{\vert \{ i \vert \exists s_{d} \in D s.t. i \subseteq s_{d} \} \vert}$ where $i$ is an item and . Moreover, the frequency of an item $i$ in the subset $S$ refers to the ratio of its occurrence times in $S$ with respect to the size of $S$, and thus \textit{average frequency of the covered items} in subset $S$ is defined as $AF(S \vert D) \equiv \frac{1}{\vert I_{S} \vert} \sum \nolimits_{i \in I_{S}} \frac{}{} $, where $I_{S} \equiv \{i \vert  \}$ 

\subsubsection{Sequence Coverage} \label{SeqCov}

\begin{figure}[!t]%[htb]	%\begin{figure*}[!t]
\centering
  \subfloat[Sequence coverage on DS1-DS3 ($k=30$)]{ \label{SeqCov1_1}
   \includegraphics[width=.23\textwidth]{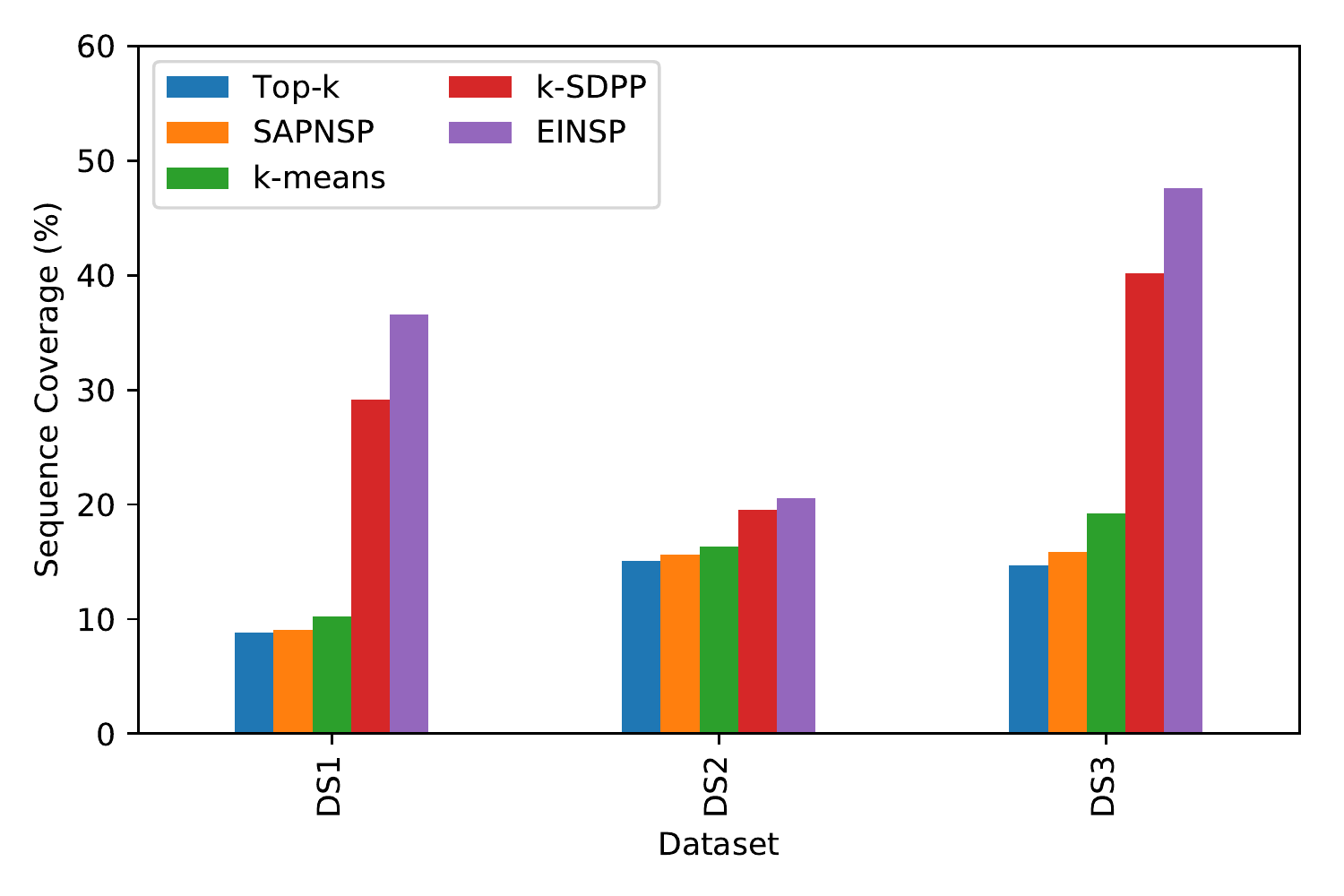}}\hfill%\\
  \subfloat[Sequence coverage on DS4-DS6 ($k=30$)]{ \label{SeqCov1_2}
   \includegraphics[width=.23\textwidth]{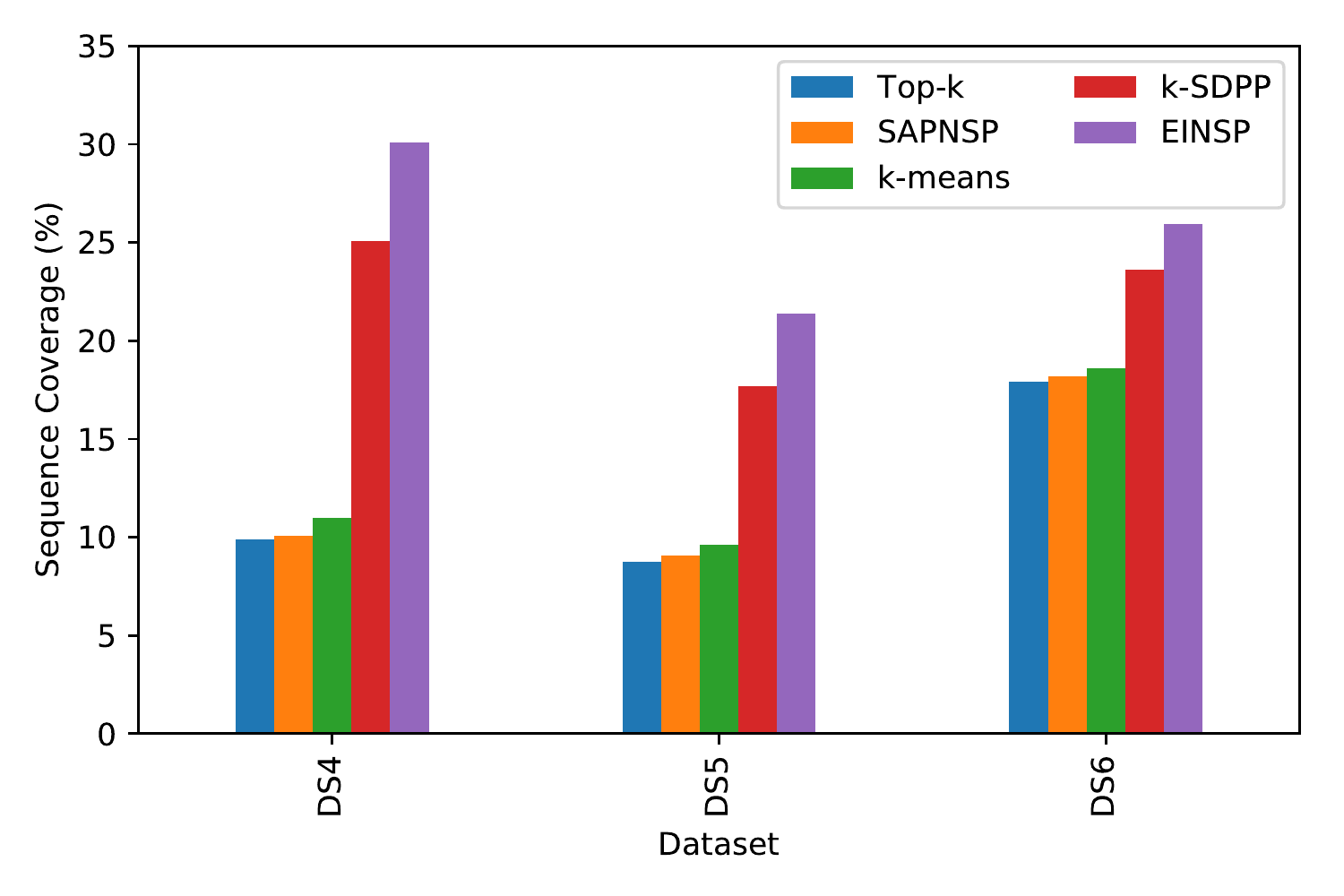}}\\ 
  \subfloat[Sequence coverage on DS1-DS3 ($k=150$)]{  \label{SeqCov1_3}
   \includegraphics[width=.23\textwidth]{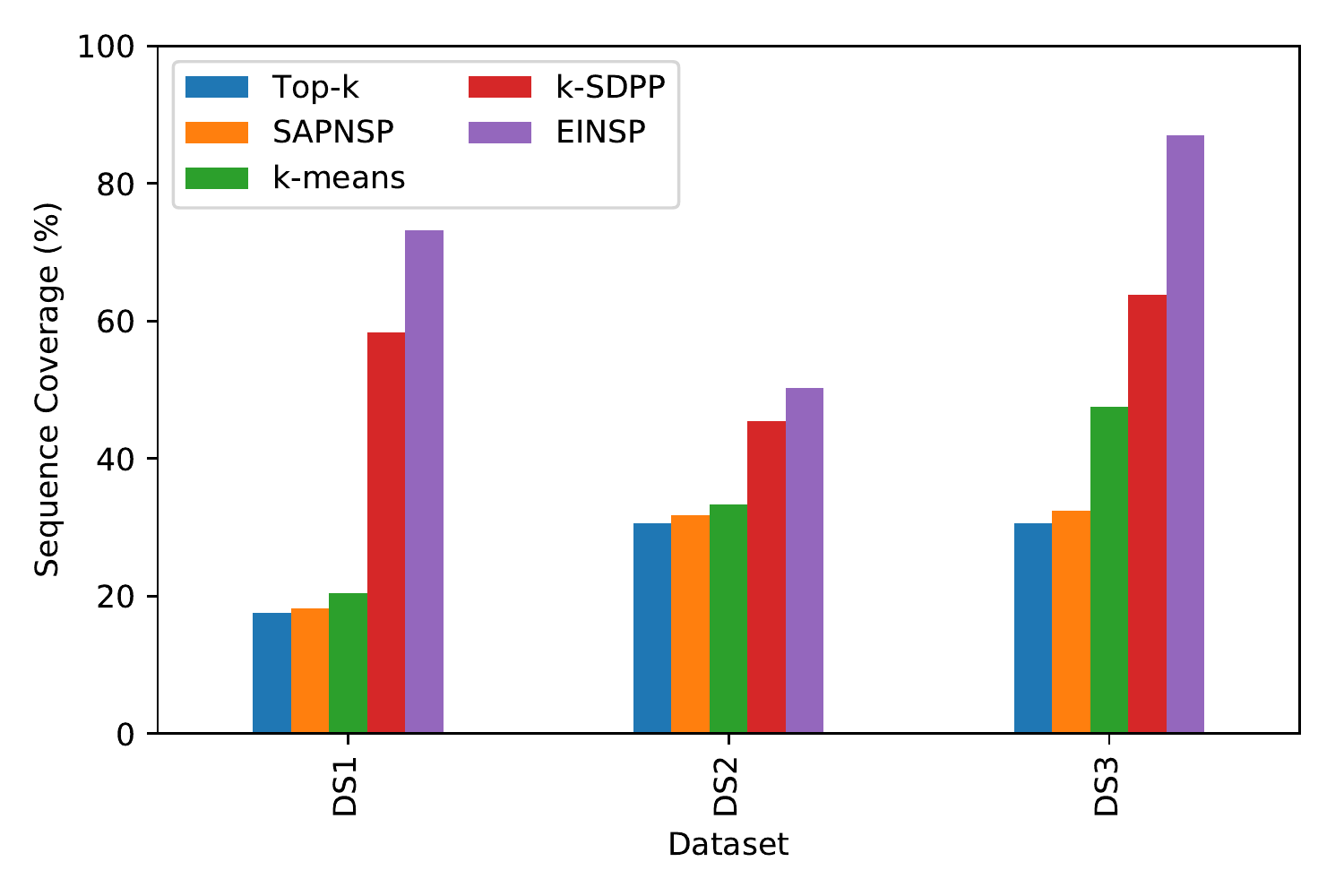}}\hfill%\\
  \subfloat[Sequence coverage on DS4-DS6 ($k=150$)]{\label{SeqCov1_4}
   \includegraphics[width=.23\textwidth]{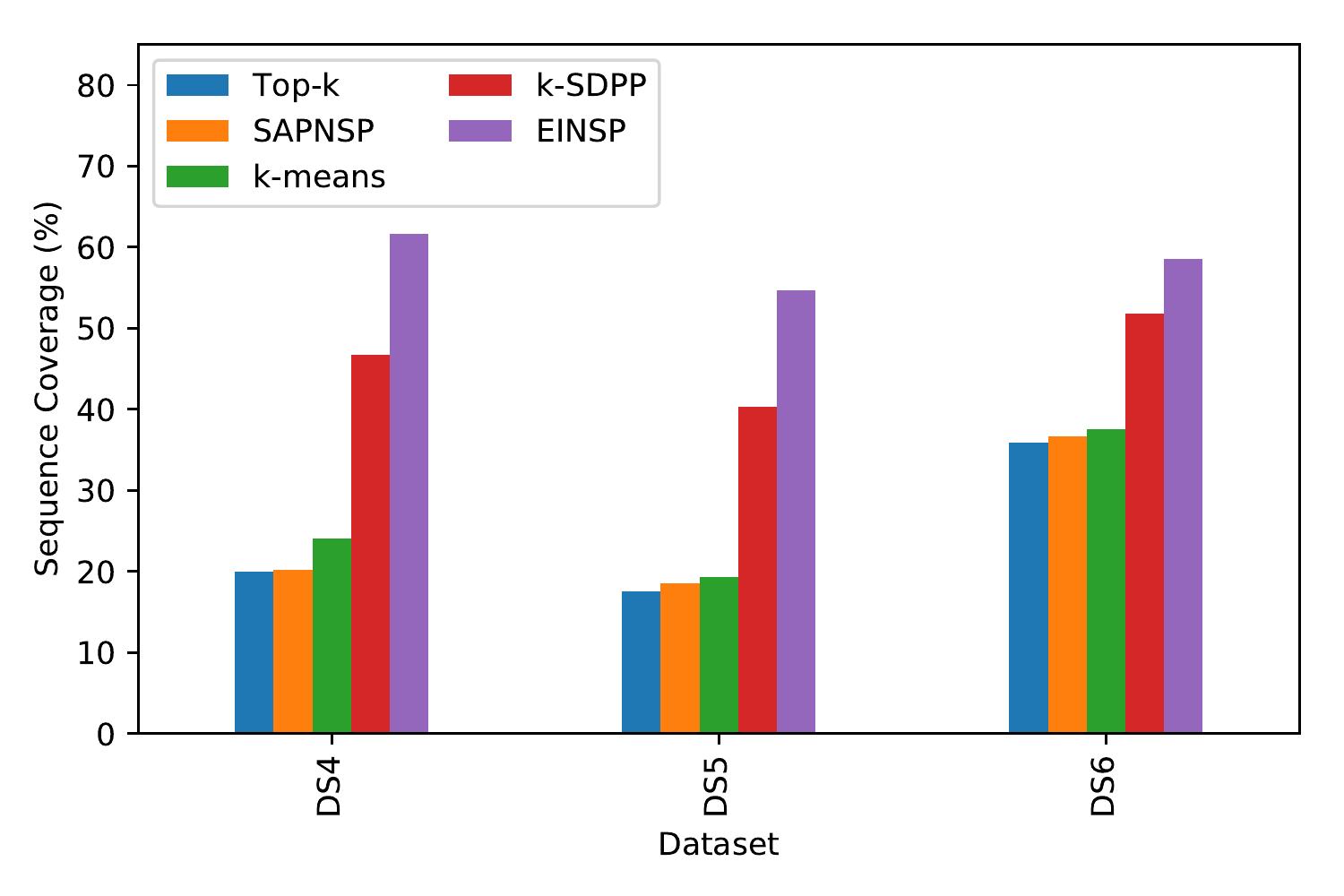}}\\ 
\caption{Sequence Coverage on Datasets DS1 to DS6}
\label{SeqCovFig}
\end{figure}

Fig. \ref{SeqCovFig} shows the results of the \textit{sequence coverage} of EINSP and the baselines on six real-life datasets. Of all the methods, \textit{Top-k selection} performs worst as it assumes that the covered sets of the selected NSP patterns are disjoint and completely neglects the diversity of the selected subset. This also shows that this method and settings are inapplicable for real-world cases. As a result, \textit{Top-k selection} achieves a relatively lower \textit{sequence coverage} on datasets with a smaller average length, such as DS1 and DS4, since a larger proportion of data sequences covers multiple short-sized and high-frequency patterns while the NSPs with relatively rare entities are ignored. 

Compared with \textit{Top-k selection}, SAPNSP achieves a slightly better \textit{sequence coverage}. SAPNSP benefits from considering both the frequency of a pattern and the internal couplings between elements of a pattern in terms of contribution. It thus keeps some long-sized patterns and achieves a higher \textit{contribution} rate, resulting in a relatively higher diversity of the selected subset. However, the superiority of SAPNSP over \textit{Top-k selection} is very limited because it assumes the contributions of selected patterns are independent of each other and ignores the \textit{contribution} ignores diversity of the selected subset. 

Further, \textit{Top-k selection} and SAPNSP lag behind the \textit{k-means} method because the \textit{k-means} method selects patterns from each cluster of the NSP collection and guarantees that the covered set of selected patterns shares a much smaller overlap between clusters. Accordingly, the \textit{k-means} method achieves a much higher \textit{sequence coverage} on sparse data. By increasing the $k$ value, its superiority becomes more obvious. For example, DS3 is a sparse dataset with 46,084 distinct items, the \textit{sequence coverage} of the \textit{k-means} method on DS3 improves by 30\% over \textit{Top-k selection} and more than 20\% over SAPNSP with $k=30$, and around 55\% over \textit{Top-k selection} and about 46\% over SAPNSP with $k=150$. These prove the importance of measuring  NSP diversity in discovering representative NSP subsets. However, the \textit{k-means} method is incapable of jointly modeling the quality and diversity of each pattern and does not capture the underlying implicit relations between NSP patterns.

\begin{figure}[!t]%[htb]	%\begin{figure*}[!t]
\centering
  \subfloat[Item Coverage on DS1-DS3 ($k=30$)]{ \label{ItemCov1_1}
   \includegraphics[width=.23\textwidth]{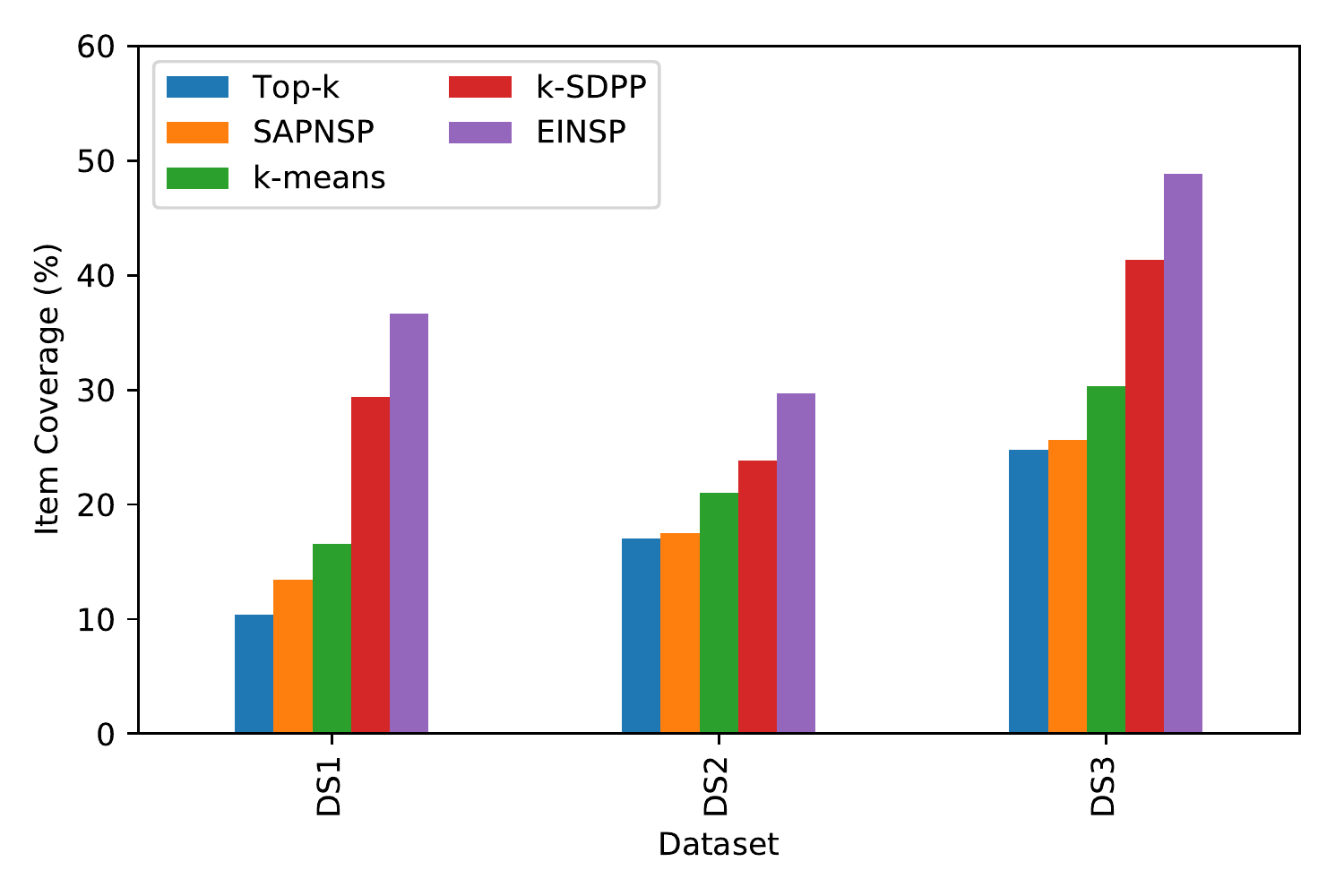}}\hfill%\\
  \subfloat[Item Coverage on DS4-DS6 ($k=30$)]{ \label{ItemCov1_2}
   \includegraphics[width=.23\textwidth]{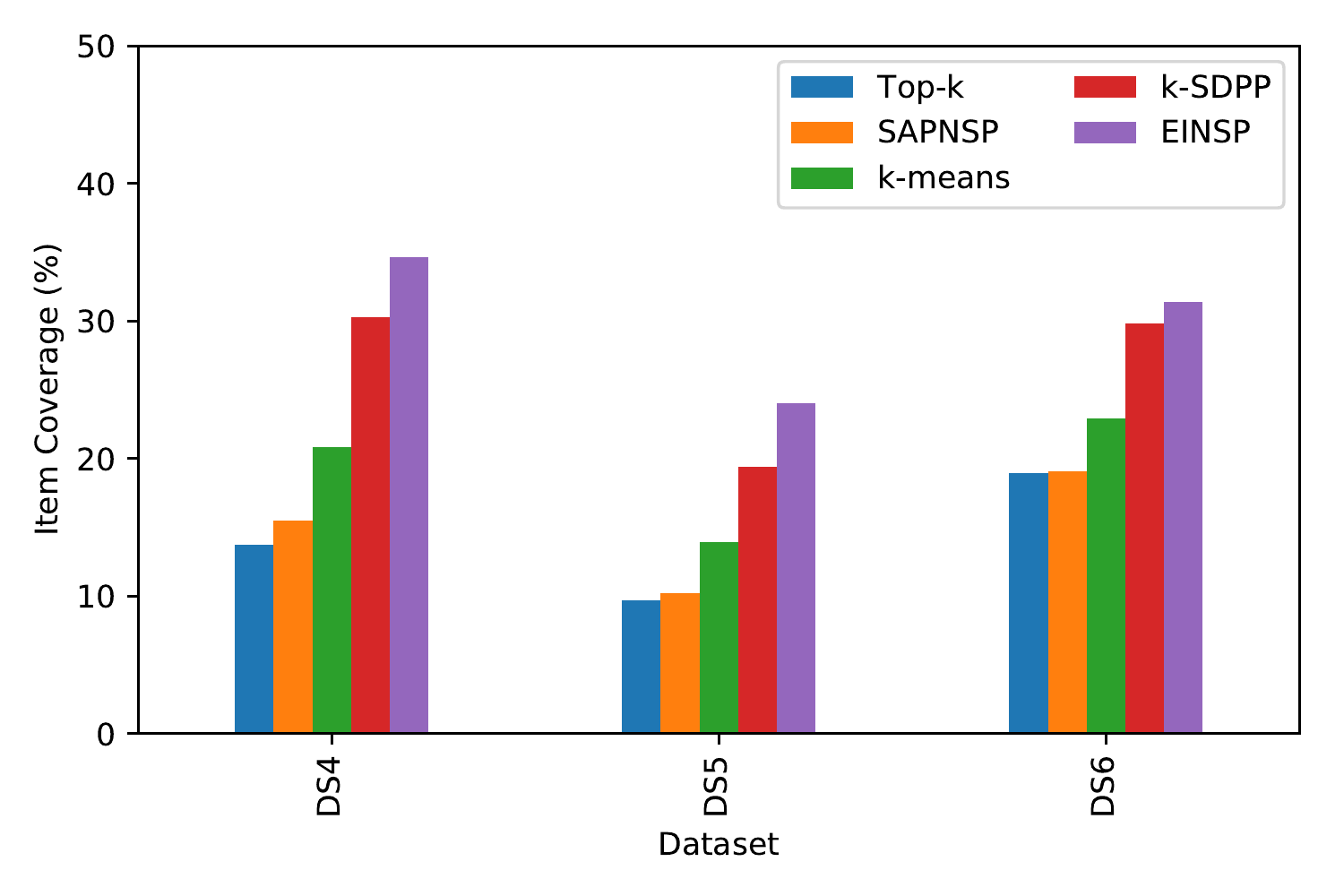}}\\ 
  \subfloat[Item Coverage on DS1-DS3 ($k=150$)]{  \label{ItemCov1_3}
   \includegraphics[width=.23\textwidth]{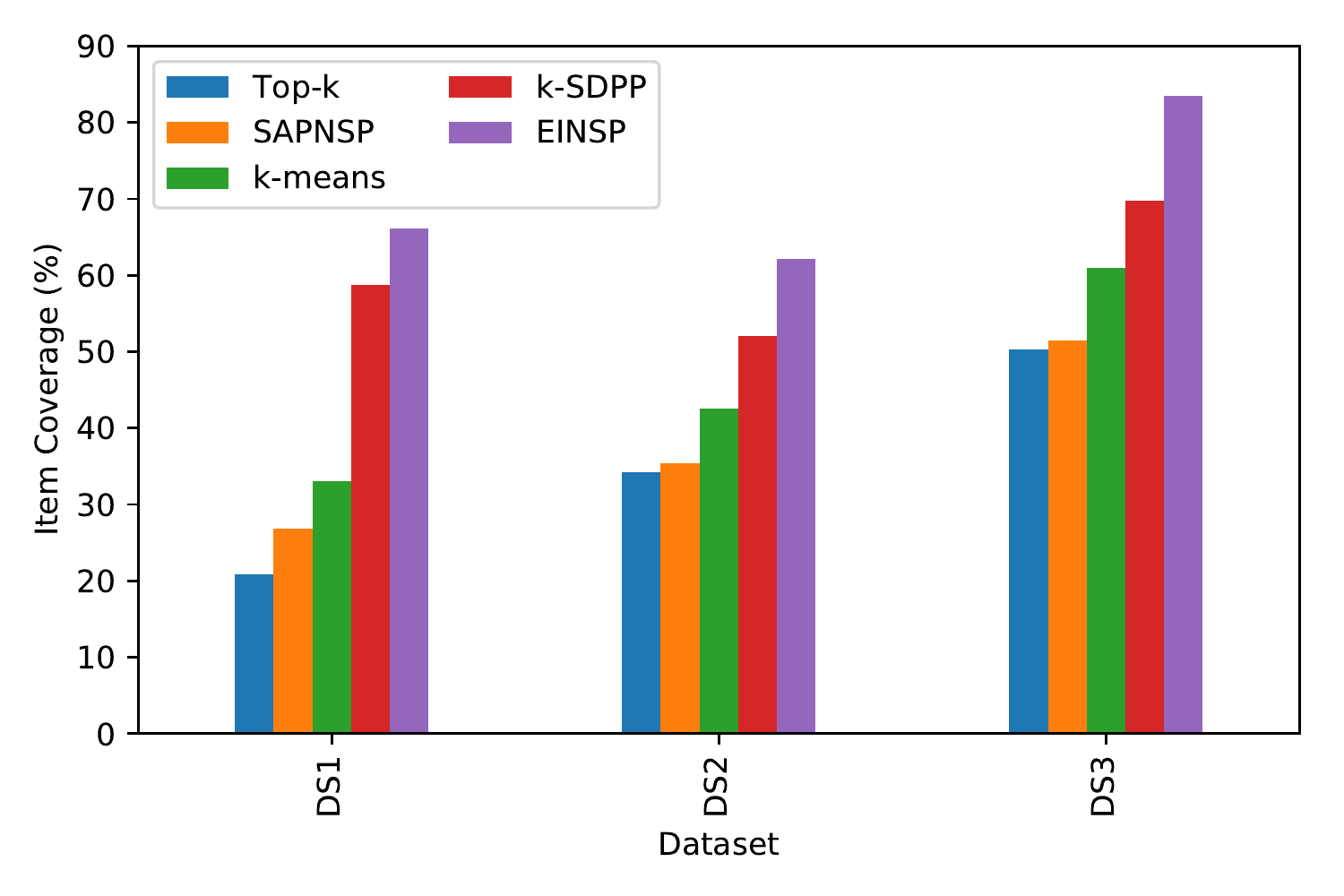}}\hfill%\\
  \subfloat[Item Coverage on DS4-DS6 ($k=150$)]{\label{ItemCov1_4}
   \includegraphics[width=.23\textwidth]{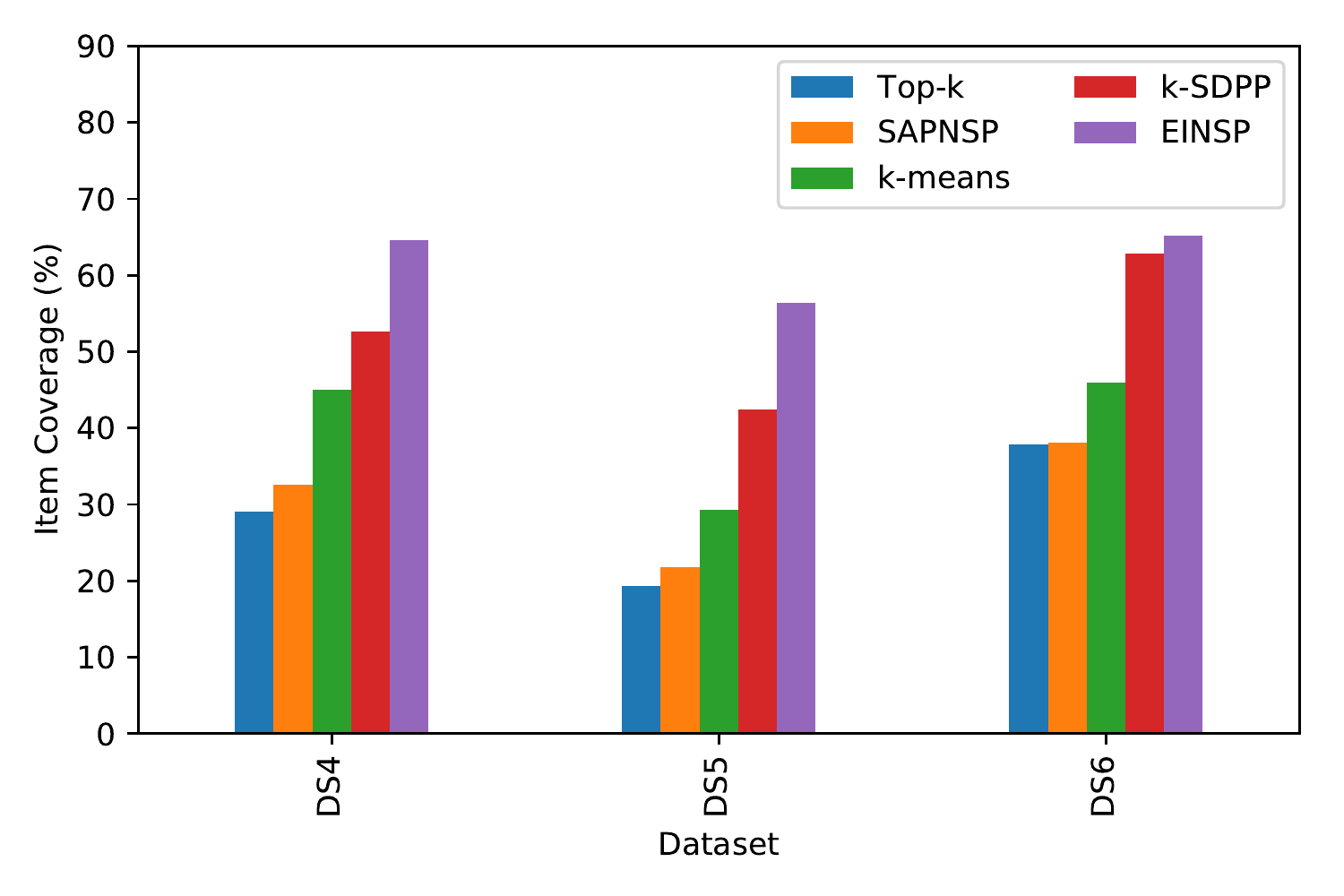}}\\ 
\caption{Item Coverage on Datasets DS1 to DS6}
\label{ItemCovFig}
\end{figure}

The experiment results indicate that EINSP and its simplified k-SDPP significantly outperform all the other baselines in terms of \textit{sequence coverage}. Particularly, EINSP achieves much higher \textit{sequence coverage} compared with the baselines by a maximum of 315\%, 36.2\%, 224\%, 204\%, 144\% and 44.9\% as well as an average of 225\%, 24.7\%, 148\%, 149\%, 106\% and 34.1\% on the datasets for $k=30$ for 30-size NSP subsets. The performance improvement made by EINSP on datasets DS2 and DS6 is less significant. When $k$ increases to 150, the improvement on these datasets increases to a maximum of 64.4\% and 63.1\% and an average of 46.1\% and 47.9\%, respectively. Overall, both EINSP and k-SDPP show strong superiority over all the baselines on six datasets, showing the effective design of modeling explicit element and pattern couplings in DPP-based NSP representations.

In addition, compared with k-SDPP which only models the explicit relations between elements and between patterns, EINSP jointly models the compound explicit and implicit element/pattern relations in NSPs. EINSP thus additionally contributes to an average of 25.4\%, 8.00\%, 27.4\%, 25.9\%, 28.3\% and 11.4\% performance improvement on the respective datasets over k-SDPP. In comparison with the \textit{k-means} method which only evaluates the importance of each pattern in each cluster in terms of pattern frequency, k-SDPP jointly considers each pattern's quality and diversity in its DPP-based design. k-SDPP thus additionally contributes to an average of 186\%, 27.9\%, 71.7\%, 111\%, 96.0\% and 32.5\% performance improvement on the six datasets over \textit{k-means} baseline. These results show that EINSP captures the NSP pattern diversity, co-occurrence-based element/pattern couplings, and non-co-occurrence-based element/pattern relations, which greatly contribute to much higher sequence coverage of the discovered NSPs. Lastly, EINSP and its simplified version k-SDPP achieve better performance and higher coverage with a higher $k$, which shows that EINSP and k-SDPP are scalable with an increase of the parameter $k$ in the DPP-based NSP graph.

In summary, our method EINSP and its simplified version k-SDPP achieve much better performance in terms of pattern coverage and diversity. The experiments demonstrate the effectiveness of EINSP's design for capturing diversified and highly probable NSPs. This is achieved by jointly modeling  co-occurrence-based explicit element and pattern couplings and implicit element and pattern relations in selecting representative NSPs. 

\subsubsection{Item Coverage} \label{ItemCov}

Fig. \ref{ItemCovFig} shows the \textit{item coverage} of the subsets selected by EINSP and baselines on the  real-life datasets. The results show that both \textit{Top-k selection} and SAPNSP achieve the worst \textit{item coverage} on all datasets and with different $k$ values. This is because they always tend to select short-size patterns consisting of only high-frequency items but ignore those rarely observed items. 

In contrast, EINSP always achieves the highest \textit{item coverage} of all the baselines on six datasets, and its selected subset covers  patterns with more distinct items. In addition, EINSP always outperforms its simplified k-SDPP by contributing to an average of 18.6\%, 22.1\%, 19.0\%, 18.6\%, 28.4\% and 4.50\%  \textit{item coverage} on the datasets. This EINSP improvement is made by introducing the implicit element/pattern relations that drive EINSP toward those non-co-occurring patterns with low-frequency item combinations. 

Moreover, with the joint consideration of both the quality and diversity of each pattern, k-SDPP beats the \textit{k-means} method by an average of 77.6\%, 17.9\%, 25.3\%, 31.0\%, 41.8\% and 33.4\% on the six datasets. This demonstrates that capturing both explicit and implicit element/pattern relations can greatly enlarge the item coverage of the selected NSP subset. This especially works for  low-frequency items which are usually discarded by frequency and downward closure property-based NSP methods such as those baselines. 

\subsubsection{Average Item Frequency} \label{AcifCov}

\begin{figure}[!t]%[htb]	%\begin{figure*}[!t]
\centering
  \subfloat[Average Item Frequency on DS1-DS3 ($k=30$)]{ \label{AcifCov1_1}
   \includegraphics[width=.23\textwidth]{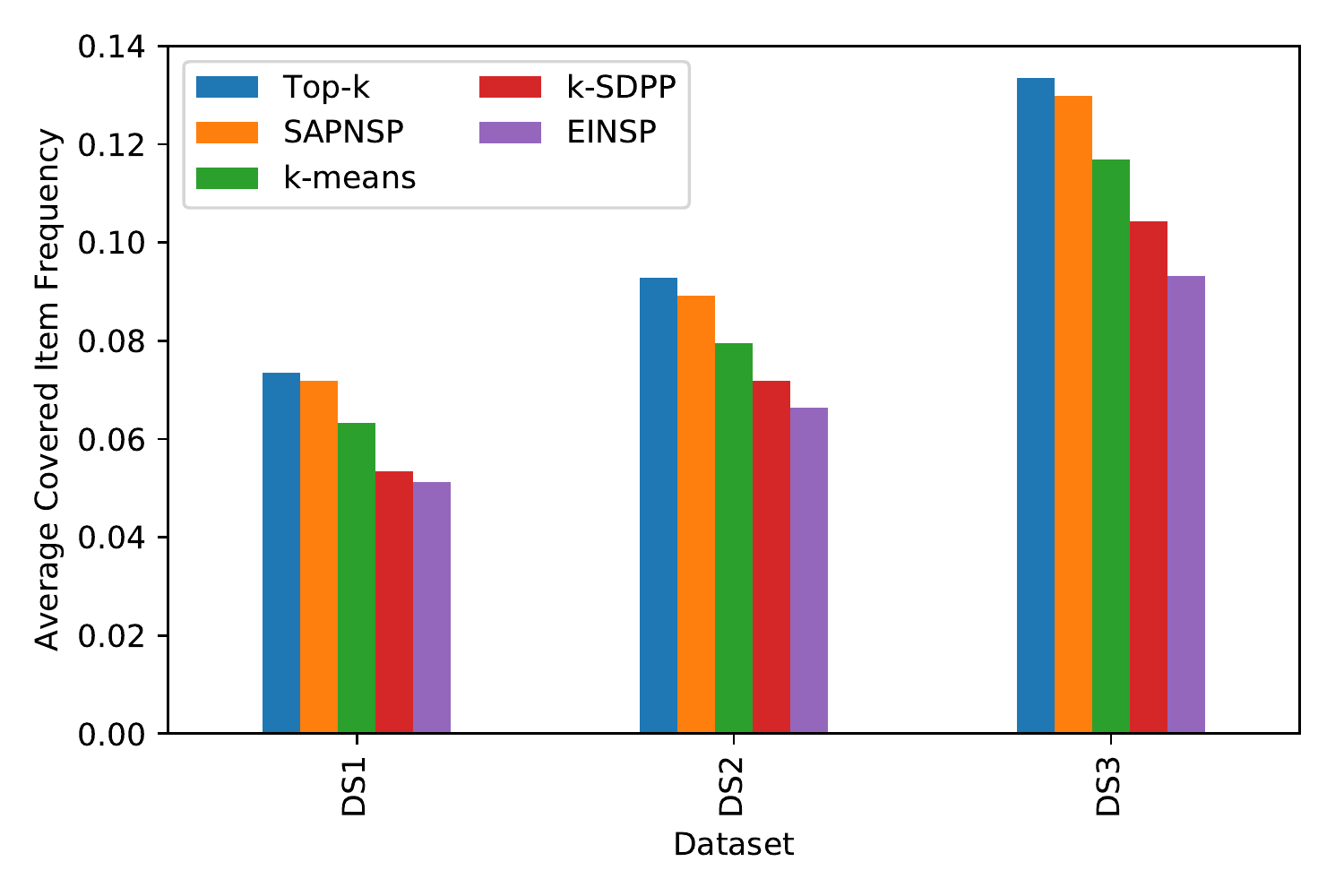}}\hfill%\\
  \subfloat[Average Item Frequency on DS4-DS6 ($k=30$)]{ \label{AcifCov1_2}
   \includegraphics[width=.23\textwidth]{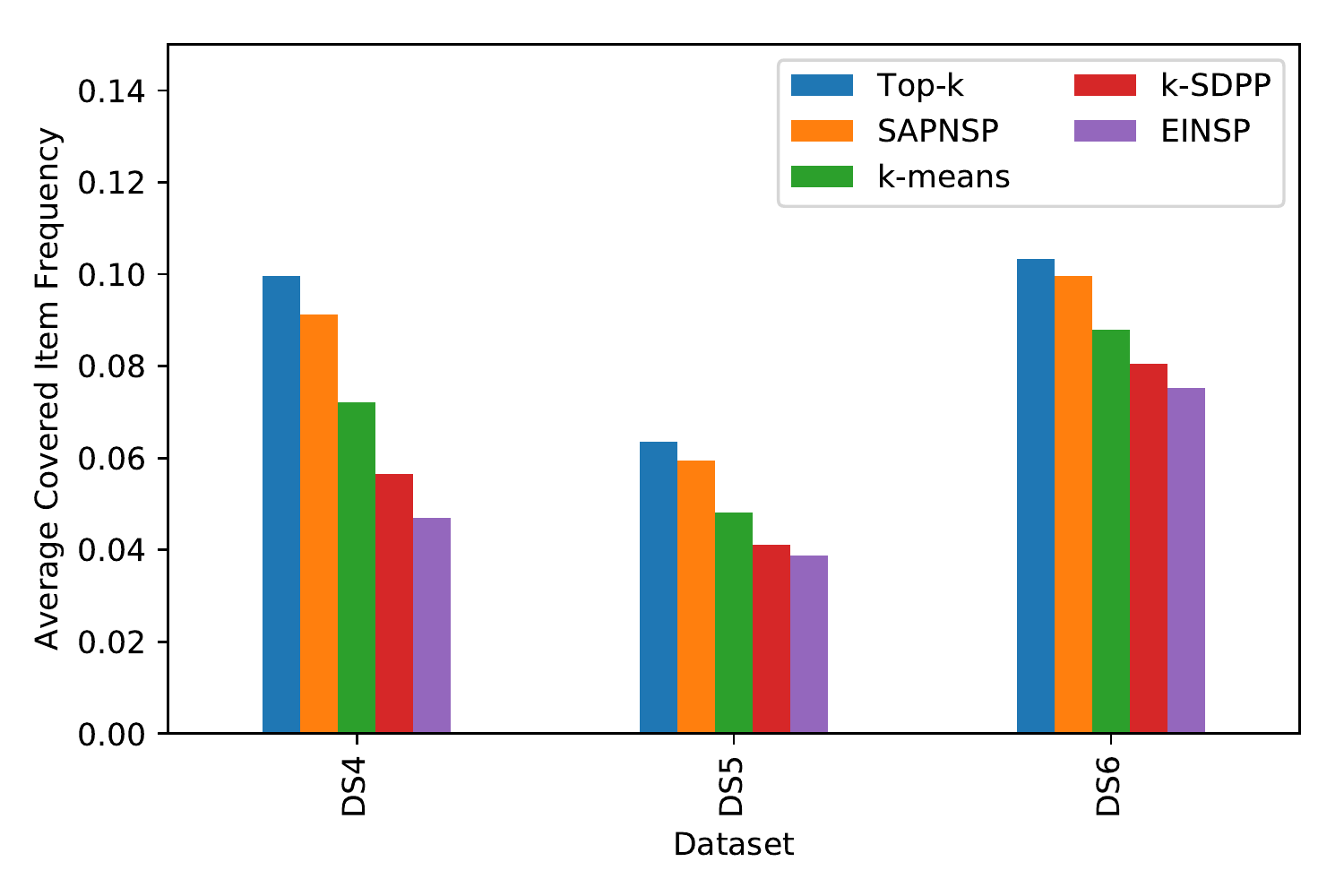}}\\ 
  \subfloat[Average Item Frequency on DS1-DS3 ($k=150$)]{  \label{AcifCov1_3}
   \includegraphics[width=.23\textwidth]{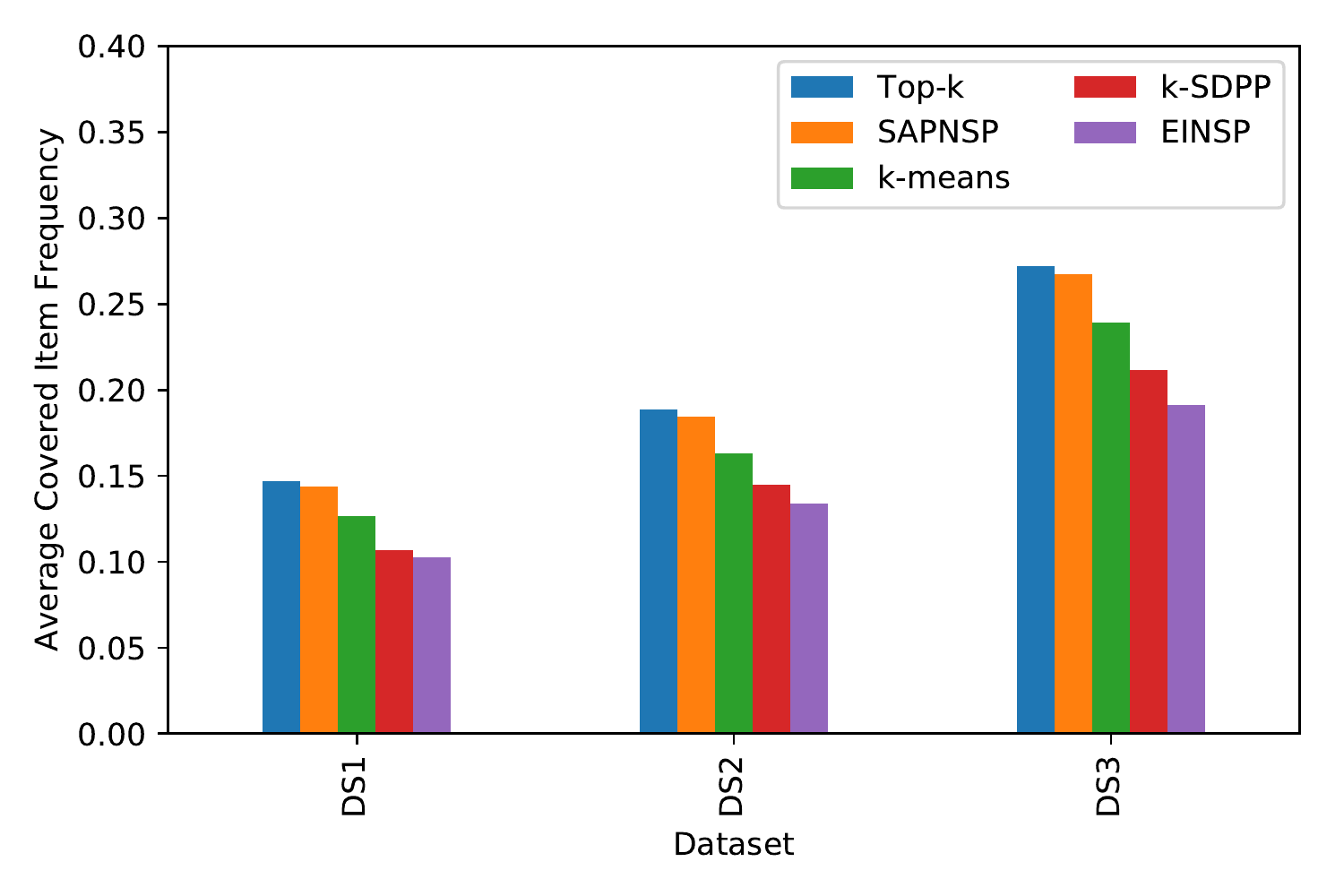}}\hfill%\\
  \subfloat[Average Item Frequency on DS4-DS6 ($k=150$)]{\label{AcifCov1_4}
   \includegraphics[width=.23\textwidth]{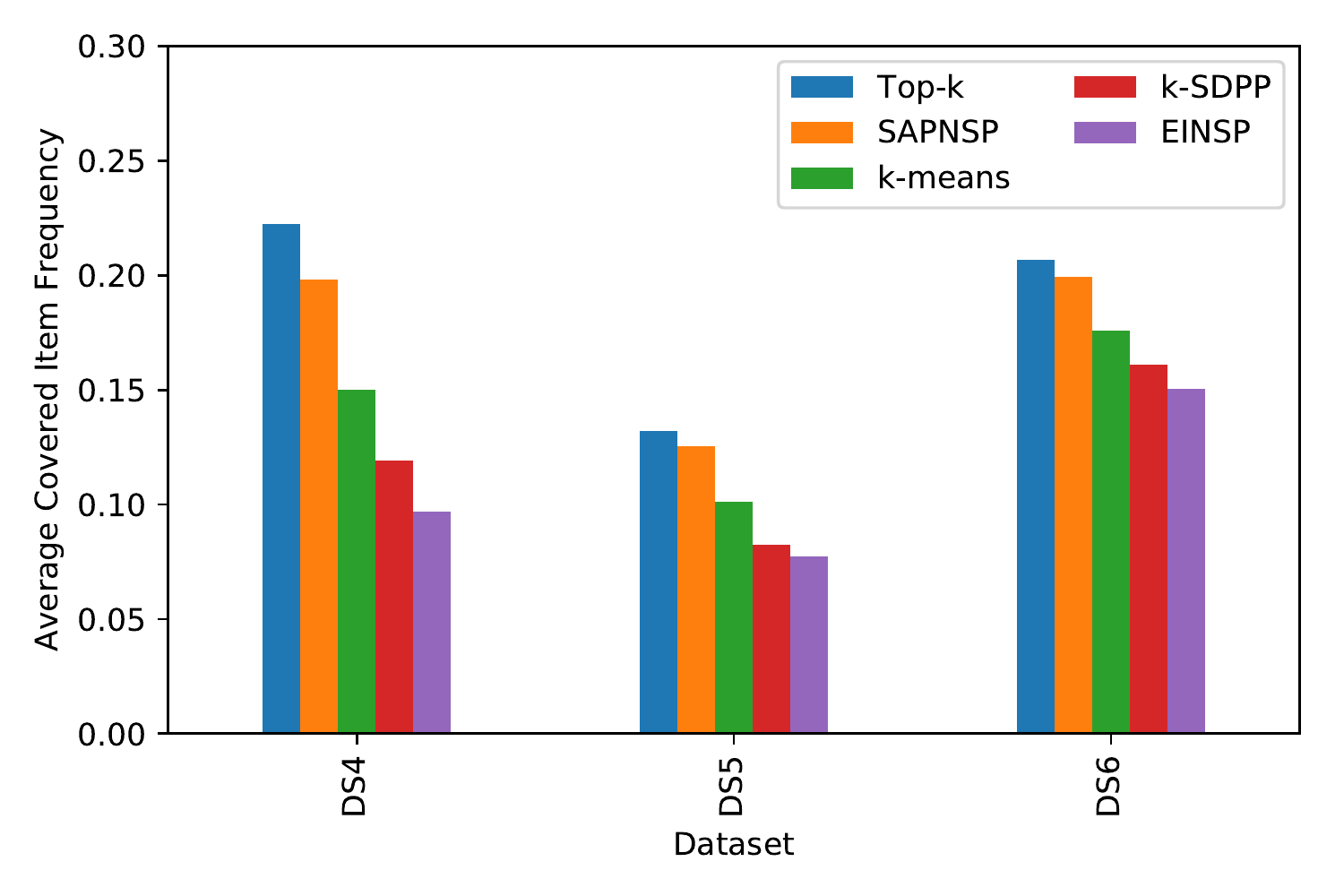}}\\ 
\caption{Average Item Frequency on Datasets DS1 to DS6}
\label{AcifCovFig}
\end{figure}

A comparison of the \textit{average item frequency} of the  items covered by EINSP and the baselines on all datasets is shown in Fig. \ref{AcifCovFig}. The \textit{average item frequency} of the items covered by the NSP subsets selected by \textit{Top-k selection} and SAPNSP is always much higher than the other three methods. This shows that both \textit{Top-k selection} and SAPNSP tend to select patterns combining a smaller number of high-frequency items due to their strong reliance on the frequency and downward closure property-based NSP selection criteria. This results in the high item frequency of selected NSP subsets. 

In comparison, all methods considering the diversity of the selected NSP subset achieve a lower \textit{average item frequency} and thus involve more rarely observed items. This may make the findings more diversified, novel and actionable. In particular, the \textit{average item frequency} of EINSP-selected NSPs is only 69.8\%, 71.1\%, 70.1\%, 45.3\%, 59.8\% and 72.8\%  of those by \textit{Top-k selection} respectively on datasets DS1 to DS6, and 81.1\%, 82.6\%, 79.8\%, 64.8\%, 78.5\% and 85.6\% of those by the \textit{k-means} method. These results show that the NSPs selected by EINSP are more balanced and diverse in terms of distinct items covered in the final NSPs. 

Combining the experimental findings in Sections \ref{ItemCov} and \ref{ItemCov}, the NSP subsets selected by EINSP not only cover a larger proportion of distinct items but also contain a higher proportion of those NSPs with relatively low-frequency items. EINSP thus produces more balanced, diversified and novel NSP items and patterns from the entire dataset. In contrast, the baseline methods suffer in these aspects from their limitation in relation to the frequentist-based design. These experiments further show (1) the importance of modeling not only direct and observable co-occurrence-based element/pattern couplings but also the indirect and implicit couplings between elements and patterns in pattern discovery \cite{Coupling2015cao}; and (2) the necessity of measuring pattern quality in terms of not only statistical significance but also pattern diversity and novelty in representative NSP discovery.

\subsection{Average Pattern Size} \label{PatSize}

\begin{figure}[!t]%[htb]	%\begin{figure*}[!t]
\centering
  \subfloat[Average Pattern Size on DS1-DS3 ($k=30$)]{ \label{PatSize1_1}
   \includegraphics[width=.23\textwidth]{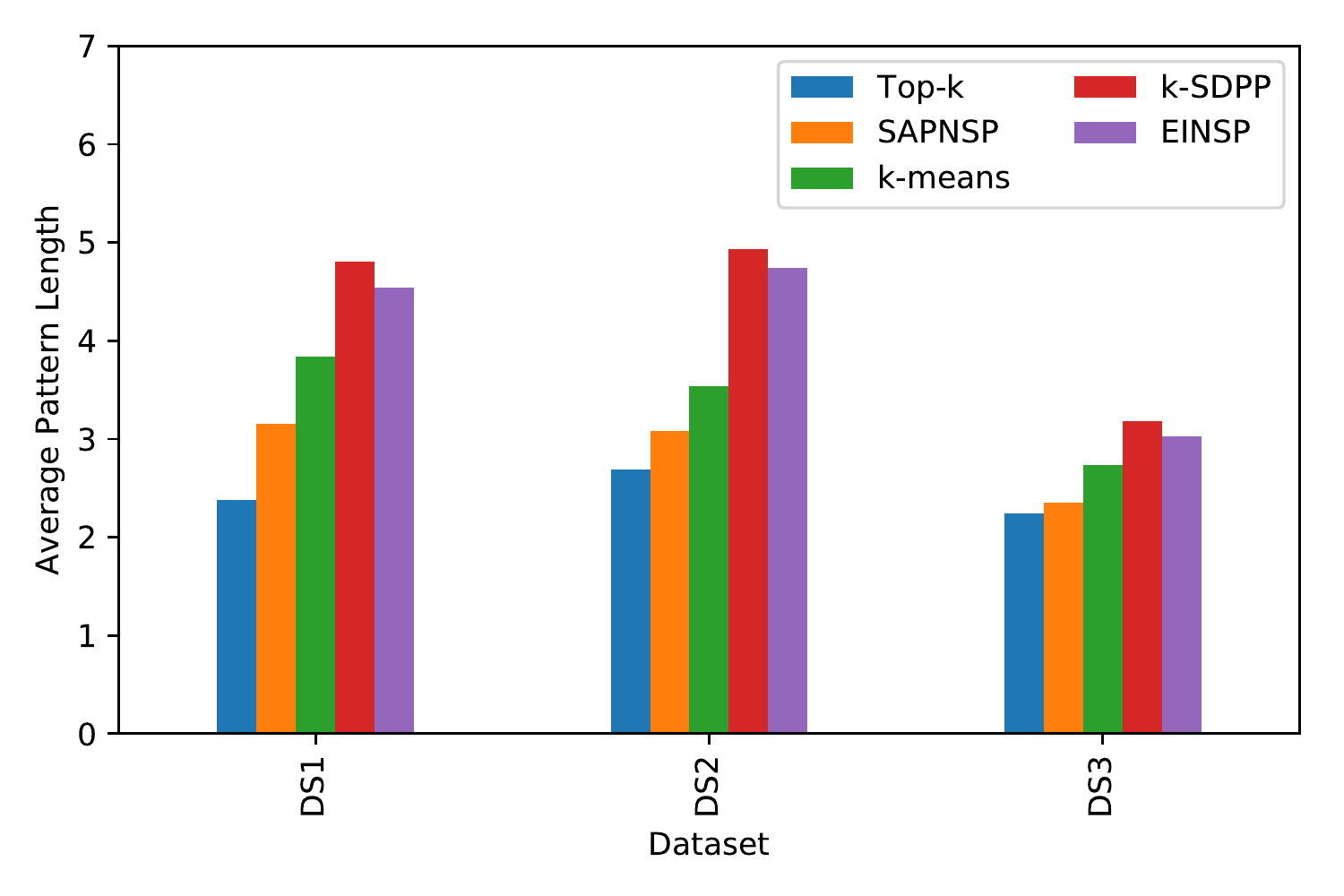}}\hfill%\\
  \subfloat[Average Pattern Size on DS4-DS6 ($k=30$)]{ \label{PatSize1_2}
   \includegraphics[width=.23\textwidth]{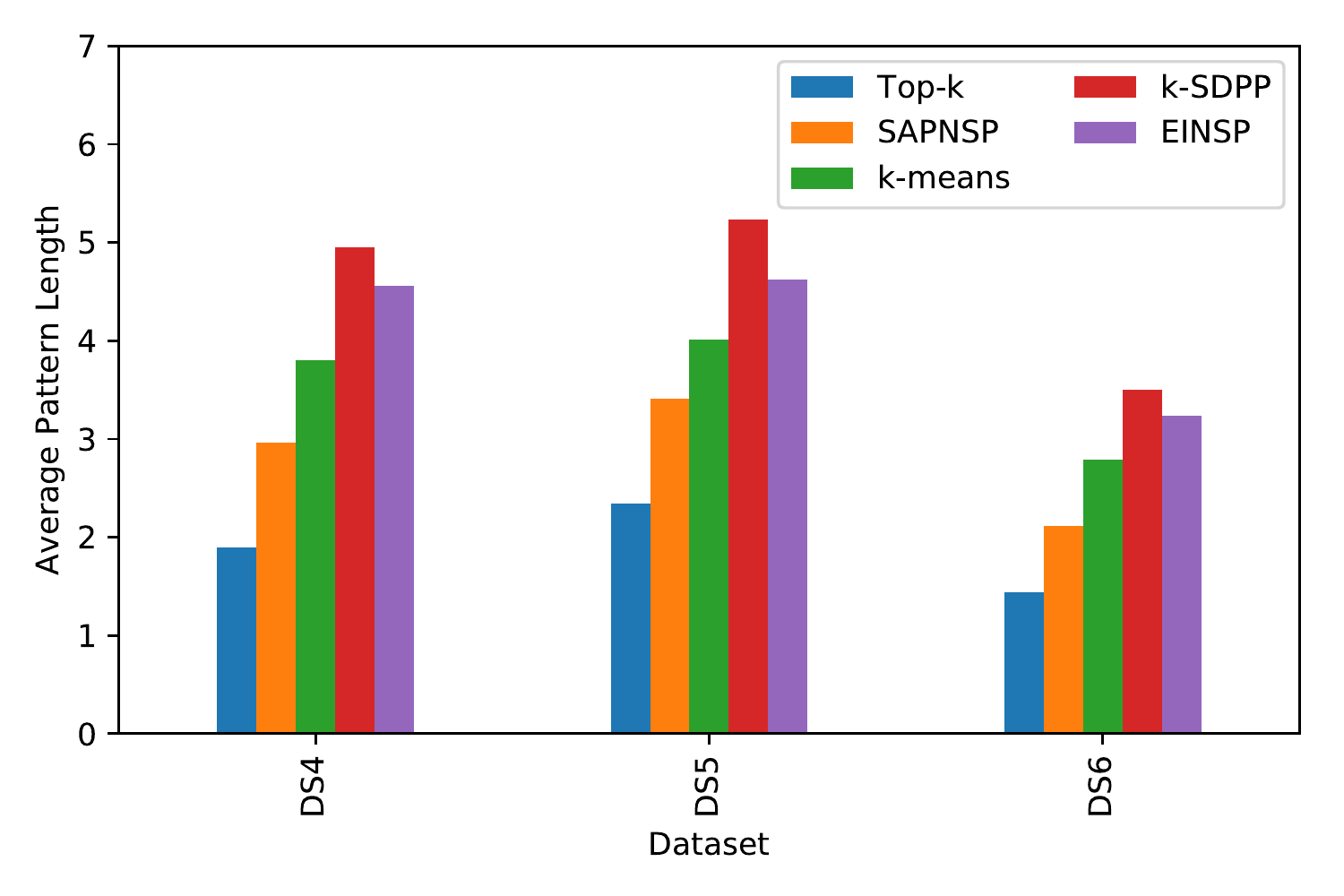}}\\ 
  \subfloat[Average Pattern Size on DS1-DS3 ($k=150$)]{\label{PatSize1_3}
   \includegraphics[width=.23\textwidth]{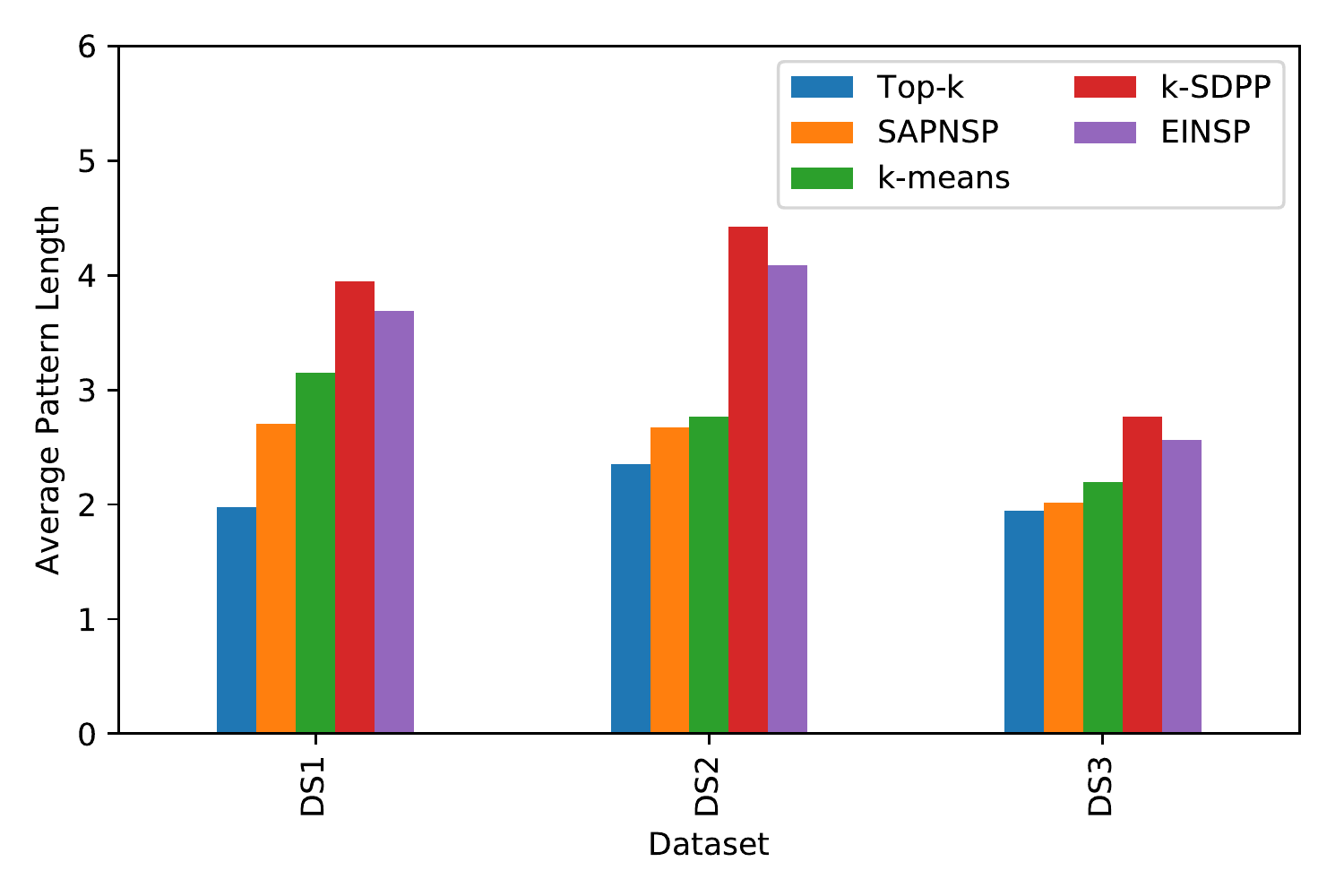}}\hfill%\\
  \subfloat[Average Pattern Size on DS4-DS6 ($k=150$)]{\label{PatSize1_4}
   \includegraphics[width=.23\textwidth]{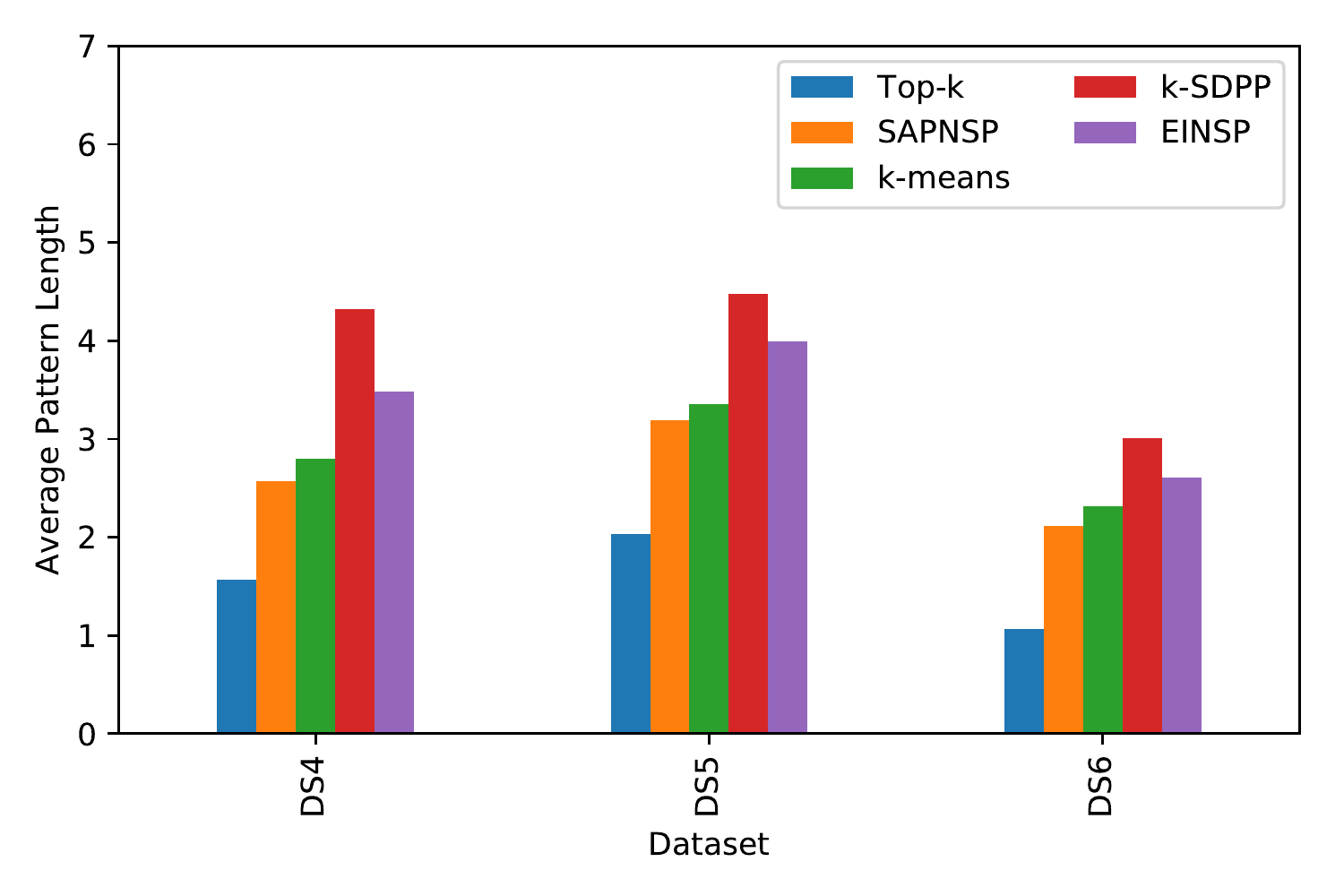}}\\ 
\caption{Average Pattern Size Comparison on DS1 to DS6}
\label{PatSizeFig}
\end{figure}

The \textit{average pattern size} of selected subsets measures the selected NSP subset quality since a longer-sized pattern captures higher long-range dependencies between the elements in the pattern. However, in  traditional frequentist-based NSP mining, long-sized patterns are more likely discarded because of their low frequency. Fig. \ref{PatSizeFig} illustrates the \textit{average pattern size} of EINSP and the baselines on six real-life datasets. First, the \textit{average pattern size} of \textit{Top-k selection} is always much smaller than that of other methods. Only a small number of its selected patterns contain more than three elements  because short-size patterns usually have higher frequency. This result is consistent with the results of item coverage in Section \ref{ItemCov}. 

Second, SAPNSP achieves a slightly bigger \textit{average pattern size} than \textit{Top-k selection}, because SAPNSP applies \textit{contribution} to pattern selection which partly favors high-frequency but short-size patterns. However, the downward closure property of \textit{contribution} indicates that a pattern always has a higher contribution than any of its super-sequences, leading to short-sized patterns being more likely selected. Accordingly, the average size of the patterns selected by SAPNSP is always much shorter than that of k-SDPP and EINSP. This shows SAPNSP cannot discover representative NSPs with long-sized patterns and diversified subsets. 

Further, the \textit{k-means} method achieves a bigger \textit{average pattern size} than the aforementioned two methods. This is because the \textit{k-means} method  groups patterns into $k$ clusters w.r.t. their diversity and thus allows those clusters with long-sized patterns. However, the \textit{k-means} method only selects the most frequent patterns in a cluster, which are the shortest in size in each belonging cluster. 

Contrary to the \textit{k-means} method, k-SDPP applies  \textit{explicit pattern quality} to evaluate the importance of an NSP pattern, which assigns a high quality to a relatively longer pattern as shown in Eq. (\ref{q_e}). Consequently, more long-sized patterns are selected, contributing to the largest \textit{average pattern size}. As a result, the \textit{average pattern size} of k-SDPP selections improves by an average of 49.2\%, 62.8\%, 36.2\%, 67.6\%, 47.0\% and 53.9\% over SAPNSP and 25.5\%, 49.6\%, 21.1\%, 42.2\%, 32.1\% and 27.8\% over \textit{k-means} on the respective datasets. 

In addition, compared with k-SDPP, EINSP achieves a smaller \textit{average pattern size} because it uses  \textit{implicit pattern quality} to select NSPs. It assigns a lower \textit{implicit relation strength} (IRS) to a longer-size pattern, similar to \cite{wang2017inferring}. Compared with the baselines, EINSP contributes to an average improvement of 40.2\%, 53.5\%, 27.9\%, 44.9\%, 30.3\% and 38.0\% over SAPNSP and 17.8\%, 40.9\%, 13.7\%, 22.3\%, 17.2\% and 14.3\% over the \textit{k-means} method w.r.t. the \textit{average pattern size} on the respective datasets. Lastly, as seen in Section \ref{PatRel}, introducing implicit pattern relations into EINSP further improves the \textit{average implicit relation strength} of the subset selected by EINSP at the expense of a smaller average pattern size. This shows EINSP selects NSPs that are significant, diverse and indirectly coupled, which cannot be identified by the existing NSP methods.

\subsection{Average Implicit Pattern Relation Strength} \label{PatRel} %\textit{average implicit relation strength}

\begin{figure}[!t]%[htb]	%\begin{figure*}[!t]
\centering
  \subfloat[Average IRS on DS1-DS3 ($k=30$)]{ \label{PatRel1_1}
   \includegraphics[width=.23\textwidth]{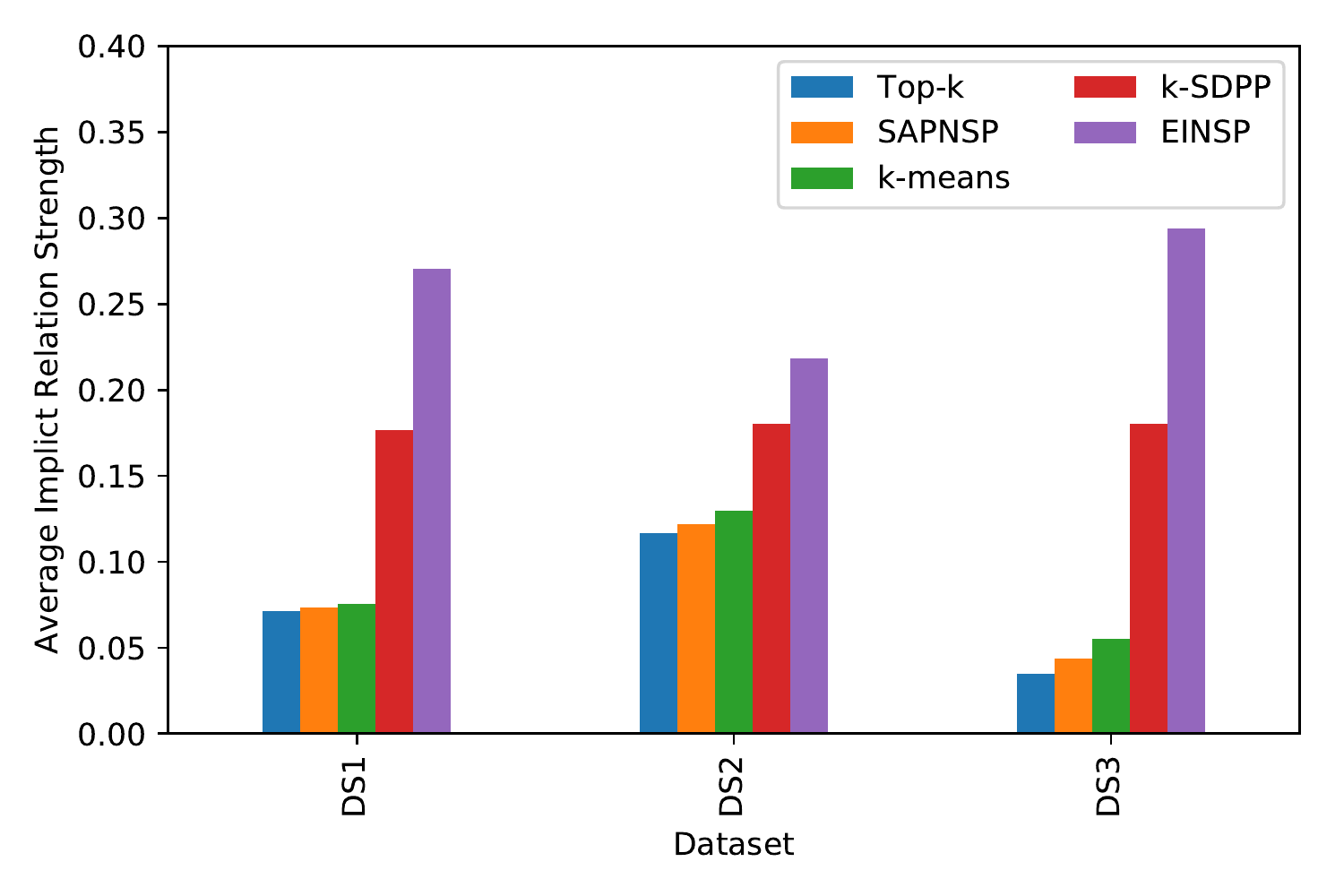}}\hfill%\\
  \subfloat[Average IRS on DS4-DS6 ($k=30$)]{ \label{PatRel1_2}
   \includegraphics[width=.23\textwidth]{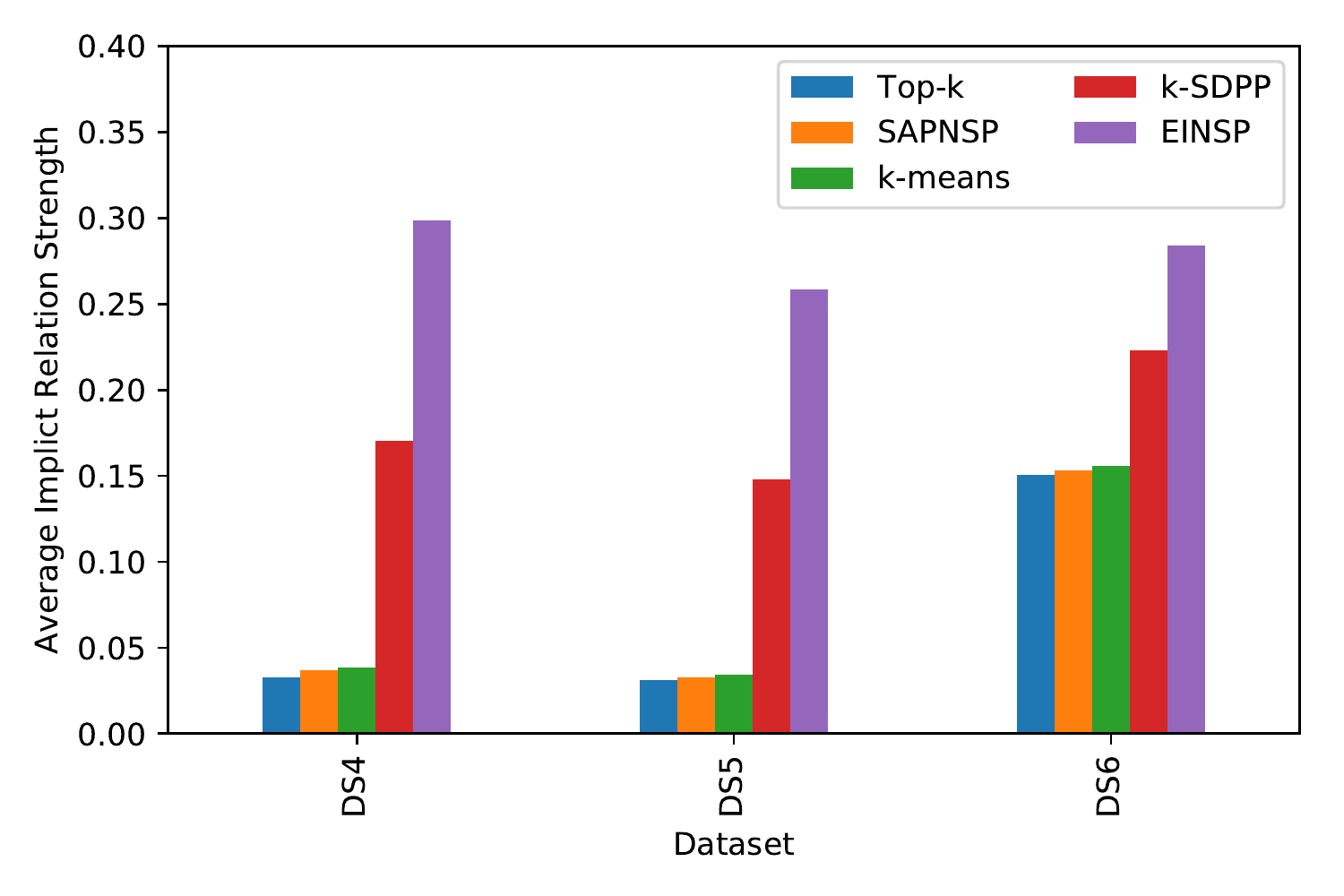}}\\ 
  \subfloat[Average IRS on DS1-DS3 ($k=150$)]{  \label{PatRel1_3}
   \includegraphics[width=.23\textwidth]{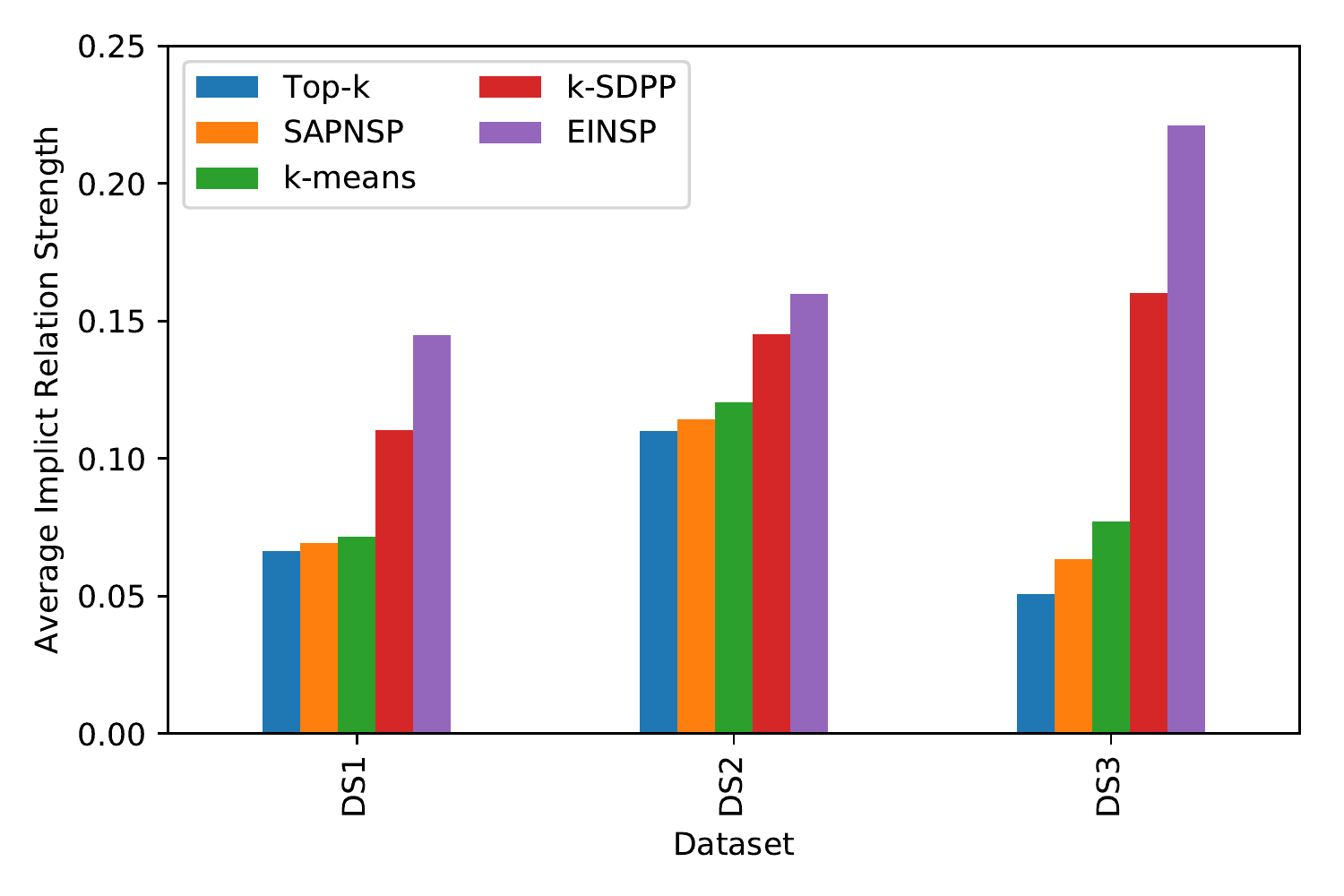}}\hfill%\\
  \subfloat[Average IRS on DS4-DS6 ($k=150$)]{\label{PatRel1_4}
   \includegraphics[width=.23\textwidth]{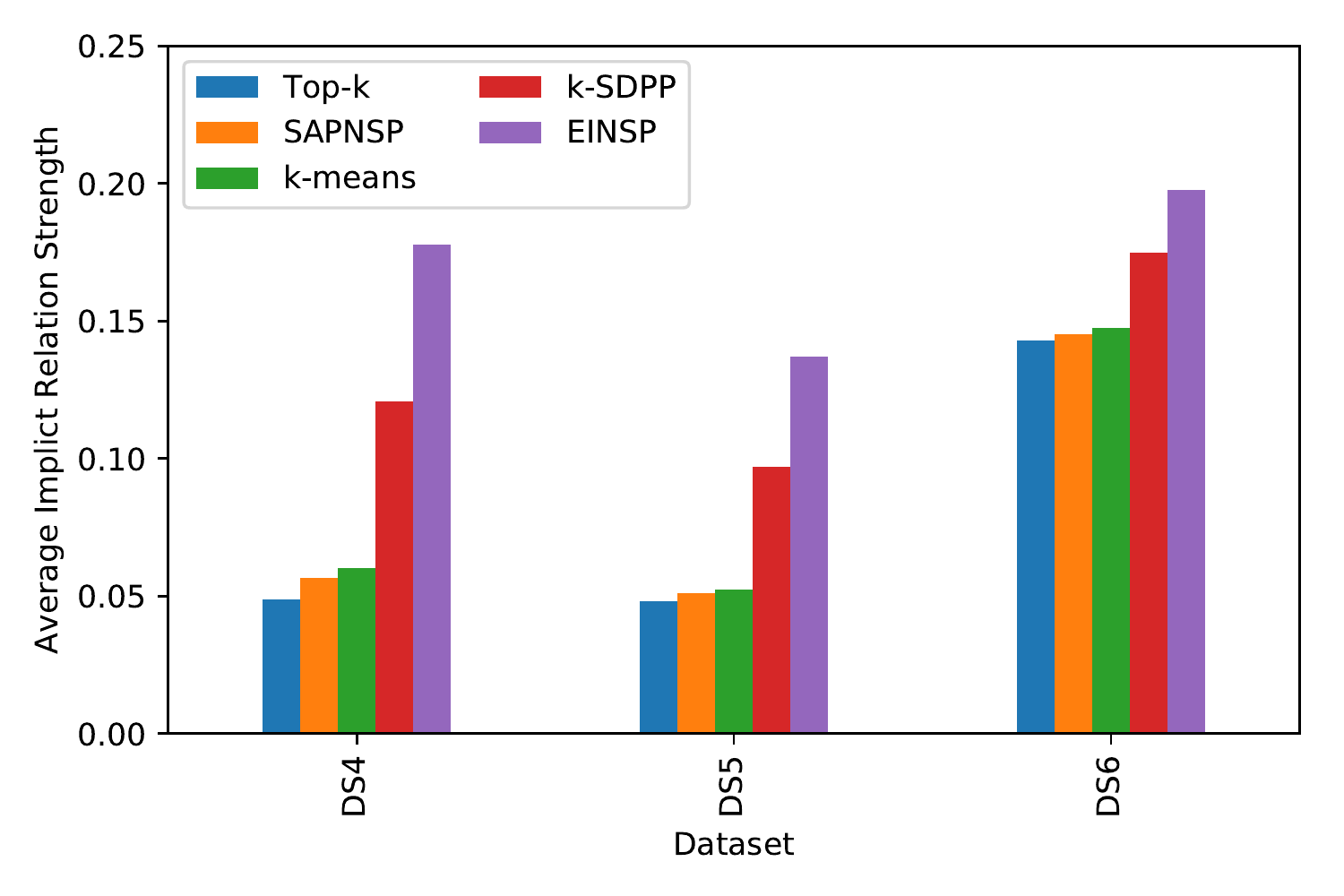}}\\ 
\caption{Average Implicit Pattern Relation Strength on DS1 to DS6}
%\caption{Average Pattern Relevance Comparison on DS1-DS8}  \textit{average implicit relation strength}
\label{PatRelFig}
\end{figure}

As discussed in Section \ref{secintro}, some lowly frequent but highly implicitly coupled patterns may reveal complex but insightful knowledge to inform business decision-making. Accordingly, the \textit{average implicit pattern relation strength} measures the quality of the subset selected by an NSP miner, with the experiment results shown in Fig. \ref{PatRelFig}. 

The three baselines \textit{Top-k selection}, SAPNSP and the \textit{k-means} always achieve similar but low \textit{average implicit pattern relation strength}. This is because they do not consider the implicit relations between patterns, i.e., their co-occurrence-based pattern selection ignores indirectly coupled elements and itemsets and their patterns. In contrast, k-SDPP achieves much higher \textit{average implicit pattern relation strength} by a maximum of 148.8\%, 418.7\%, 419.2\% and 375.5\% for $k=30$ and an average of 141.2\%, 320.0\%, 374.4\% and 354.2\% for $k=150$ on datasets DS3, DS5, DS6 and DS7, respectively. Accordingly, k-SDPP shows an improvement over the \textit{average implicit pattern relation strength} by a maximum of 66.4\%, 214.6\%, 148.0\% and 101.9\% and an average of 60.1\%, 158.0\%, 120.7\% and 92.0\% on these datasets. These results show that k-SDPP partially involves the indirect couplings between elements and selects those patterns consisting of non-co-occurrence-based elements.  

Benefiting from jointly modelling the explicit and implicit relations between elements and between patterns, EINSP achieves a much higher \textit{average implicit pattern relation strength} than k-SDPP by an additional 52.8\%, 21.1\%, 63.0\%, 75.3\%, 74.4\% and 27.3\% performance improvement on the six datasets with $k=30$ and 31.4\%, 10.1\%, 38.1\%, 47.4\%, 41.2\% and 13.1\% with $k=150$. This demonstrates the importance of modelling implicit quality and the diversity of non-co-occurring NSP elements and patterns. EINSP thus discovers a highly implicitly related subset. Generally speaking, both k-SDPP and EINSP achieve better performance for a larger subset size $k$ in terms of the \textit{average implicit pattern relation strength} on all datasets but show greater superiority over the baselines for a smaller subset size.

In summary, EINSP achieves significantly bigger \textit{pattern coverage}, bigger \textit{average pattern size}, and higher \textit{average implicit pattern relation strength} of its selected NSP subsets than all of the baselines. The aforementioned experiments conclude the necessity and effectiveness of jointly modelling explicit and implicit element/pattern relations in discovering statistically significant, diverse, and indirectly coupled NSP pattern sets. Further, in Section \ref{MetSen}, we  show the scalability of EINSP  in terms of different data factors.

\begin{table*}[!t]
\renewcommand{\arraystretch}{1.3}
\caption{Sequence Coverage Sensitivity of Methods against Data Factors on $k=30$}
\label{MS1}
\centering
\scalebox{0.85}[0.85]{  %\scalebox{\textwidth}{!}{
\begin{tabular}{| p{25pt} | p{100pt} |  p{30pt} | p{30pt} | p{30pt} | p{30pt} | p{30pt} |}
%\begin{tabular}{| p{25pt} | p{49pt} | p{15pt} | p{24pt} | p{24pt} | p{15pt} | p{15pt} |}
\hline
 Factors &  Dataset Name &  Top-k   & SAPNSP    & k-means & k-SDPP &   EINSP  \\ \hline

\multirow{5}{*}{C} & \textbf{C6}\_T6\_S8\_I8\_DB10k\_N0.1k & 14.86\% & 15.92\% & 16.90\% &  23.53\% &  32.58\% \\ \cline{2-7}
                       & \textbf{C8}\_T6\_S8\_I8\_DB10k\_N0.1k & 15.99\% &  16.65\% &  17.88\% &  24.84\% &  33.67\% \\ \cline{2-7}
                       & \textbf{C10}\_T6\_S8\_I8\_DB10k\_N0.1k & 18.52\% &  19.25\% &  20.19\% &  27.95\% &   38.12\% \\\cline{2-7}
                       & \textbf{C12}\_T6\_S8\_I8\_DB10k\_N0.1k & 20.79\% &  21.56\% &  23.11\% &  31.78\% &  42.96\% \\\cline{2-7}
                       & \textbf{C14}\_T6\_S8\_I8\_DB10k\_N0.1k & 27.37\% &  28.39\% &  29.59\% &  39.69\% &  54.78\% \\\hline
\multirow{5}{*}{T} & C10\_\textbf{T4}\_S8\_I8\_DB10k\_N0.1k & 14.39\% &  15.01\% &  15.74\% &  21.72\% &  30.05\% \\ \cline{2-7}
                       &  C10\_\textbf{T6}\_S8\_I8\_DB10k\_N0.1k & 18.52\% & 19.25\% & 20.19\% & 27.95\% & 38.12\% \\ \cline{2-7}
                       & C10\_\textbf{T8}\_S8\_I8\_DB10k\_N0.1k & 25.85\% & 26.69\% & 27.56\% & 37.86\% & 50.44\% \\\cline{2-7}
                       & C10\_\textbf{T10}\_S8\_I8\_DB10k\_N0.1k & 28.49\% & 29.50\% & 30.85\% & 42.81\% & 56.71\%  \\\cline{2-7}
                       & C10\_\textbf{T12}\_S8\_I8\_DB10k\_N0.1k & 37.19\% & 38.14\%  & 39.76\% & 53.93\% & 68.25\% \\\hline
\multirow{5}{*}{DB} & C10\_T6\_S8\_I8\_\textbf{DB10k}\_N0.1k & 18.52\% & 19.25\%  & 20.19\% & 27.95\% & 38.12\%  \\ \cline{2-7}
                       &  C10\_T6\_S8\_I8\_\textbf{DB20k}\_N0.1k & 18.41\%  & 19.17\% & 19.86\% & 27.68\% & 37.66\%  \\ \cline{2-7}
                       & C10\_T6\_S8\_I8\_\textbf{DB30k}\_N0.1k & 18.23\% & 18.96\% & 20.18\% & 28.04\% & 38.09\%  \\\cline{2-7}
                       & C10\_T6\_S8\_I8\_\textbf{DB40k}\_N0.1k & 18.04\% & 18.66\% & 19.94\% & 27.76\% & 37.57\%   \\\cline{2-7}
                       & C10\_T6\_S8\_I8\_\textbf{DB50k}\_N0.1k & 17.63\% & 18.36\% & 19.90\% & 27.72\% & 37.80\% \\\hline
\multirow{5}{*}{N} & C10\_T6\_S8\_I8\_DB10k\_\textbf{N0.1k} & 18.52\% & 19.25\% & 20.19\% & 27.95\% & 38.12\%  \\ \cline{2-7}
                       &  C10\_T6\_S8\_I8\_DB10k\_\textbf{N0.2k} & 15.17\% & 16.14\% & 16.51\% & 21.66\% & 35.15\%  \\ \cline{2-7}
                       & C10\_T6\_S8\_I8\_DB10k\_\textbf{N0.3k} & 14.51\% & 15.08\% & 15.66\% & 21.94\% & 32.62\% \\\cline{2-7}
                       & C10\_T6\_S8\_I8\_DB10k\_\textbf{N0.4k} & 13.89\% & 15.01\% & 16.05\% & 21.36\% & 32.10\%   \\\cline{2-7}
                       & C10\_T6\_S8\_I8\_DB10k\_\textbf{N0.5k} & 11.29\% & 11.95\% & 12.66\% & 17.28\% & 31.63\% \\\hline
%\hline
\hline
\end{tabular}
}
\end{table*}

\begin{table*}[!t]
\renewcommand{\arraystretch}{1.3}
\caption{Sequence Coverage Sensitivity of EINSP and Baselines w.r.t. Data Factors for $k=150$}
\label{MS2}
\centering
\scalebox{0.85}[0.85]{  %\scalebox{\textwidth}{!}{
\begin{tabular}{| p{25pt} | p{100pt} |  p{30pt} | p{30pt} | p{30pt} | p{30pt} | p{30pt} |}
%\begin{tabular}{| p{25pt} | p{49pt} | p{15pt} | p{24pt} | p{24pt} | p{15pt} | p{15pt} |}
\hline
 Factors &  Dataset Name &  Top-k   & SAPNSP    & k-means & k-SDPP &   EINSP  \\ \hline

\multirow{5}{*}{C} & \textbf{C6}\_T6\_S8\_I8\_DB10k\_N0.1k & 22.80\%  & 24.25\%  & 25.74\% & 35.41\% & 43.77\%  \\ \cline{2-7}
                       & \textbf{C8}\_T6\_S8\_I8\_DB10k\_N0.1k & 26.95\% & 28.10\%  & 29.81\% & 40.98\% & 49.25\%  \\ \cline{2-7}
                       & \textbf{C10}\_T6\_S8\_I8\_DB10k\_N0.1k & 28.67\% & 29.74\% & 31.05\% & 42.04\% & 52.09\% \\\cline{2-7}
                       & \textbf{C12}\_T6\_S8\_I8\_DB10k\_N0.1k & 33.66\% & 34.74\% & 36.95\% & 49.97\% & 62.33\%  \\\cline{2-7}
                       & \textbf{C14}\_T6\_S8\_I8\_DB10k\_N0.1k & 44.71\% & 46.34\% & 48.16\% & 63.97\% & 77.61\%  \\\hline
\multirow{5}{*}{T} & C10\_\textbf{T4}\_S8\_I8\_DB10k\_N0.1k & 24.53\%  & 25.39\% & 26.67\% & 35.96\% & 44.04\%  \\ \cline{2-7}
                       &  C10\_\textbf{T6}\_S8\_I8\_DB10k\_N0.1k & 28.67\% & 29.74\% & 31.05\% & 42.04\% & 52.09\%  \\ \cline{2-7}
                       & C10\_\textbf{T8}\_S8\_I8\_DB10k\_N0.1k & 37.50\% & 38.66\% & 39.85\% & 53.68\% & 64.68\%  \\\cline{2-7}
                       & C10\_\textbf{T10}\_S8\_I8\_DB10k\_N0.1k & 39.18\% & 40.34\% & 42.29\%  & 57.30\%  & 70.94\%  \\\cline{2-7}
                       & C10\_\textbf{T12}\_S8\_I8\_DB10k\_N0.1k & 51.39\% & 52.54\% & 54.49\% & 73.01\% & 85.40\% \\\hline
\multirow{5}{*}{DB} & C10\_T6\_S8\_I8\_\textbf{DB10k}\_N0.1k & 28.67\% & 29.74\% & 31.05\% & 42.04\% & 52.09\%  \\ \cline{2-7}
                       &  C10\_T6\_S8\_I8\_\textbf{DB20k}\_N0.1k & 30.19\% & 29.07\% & 30.95\% & 41.74\% & 51.93\%   \\ \cline{2-7}
                       & C10\_T6\_S8\_I8\_\textbf{DB30k}\_N0.1k & 30.21\% & 29.94\% & 31.08\% & 41.93\% & 52.36\%   \\\cline{2-7}
                       & C10\_T6\_S8\_I8\_\textbf{DB40k}\_N0.1k & 30.25\% & 29.40\% & 31.10\% & 41.47\% & 52.12\%    \\\cline{2-7}
                       & C10\_T6\_S8\_I8\_\textbf{DB50k}\_N0.1k & 30.26\% & 29.06\% & 31.12\% & 41.46\% & 51.73\%  \\\hline
\multirow{5}{*}{N} & C10\_T6\_S8\_I8\_DB10k\_\textbf{N0.1k} & 28.67\% & 29.74\% & 31.05\% & 42.04\% & 52.09\%  \\ \cline{2-7}
                       &  C10\_T6\_S8\_I8\_DB10k\_\textbf{N0.2k} & 23.57\% & 24.94\% & 25.85\% & 33.71\% & 45.74\%  \\ \cline{2-7}
                       & C10\_T6\_S8\_I8\_DB10k\_\textbf{N0.3k} & 22.54\% & 23.24\% & 24.10\% & 32.79\% & 41.87\% \\\cline{2-7}
                       & C10\_T6\_S8\_I8\_DB10k\_\textbf{N0.4k} & 21.46\% & 22.77\% & 24.59\% & 32.84\% & 39.50\%    \\\cline{2-7}
                       & C10\_T6\_S8\_I8\_DB10k\_\textbf{N0.5k} & 17.46\% & 18.39\% & 19.28\% & 26.60\%  & 34.06\% \\\hline
%\hline
\hline
\end{tabular}
}
\end{table*}

\subsection{Data Factor Sensitivity} \label{MetSen}

As shown in \cite{WangC19,e-NSP2}, pattern mining methods including NSP miners are often sensitive to data characteristics which are quantifiable by data factors, conforming to the results in Sections \ref{PatCov}, \ref{PatSize} and \ref{PatRel} for actionable NSP discovery. A reliable NSP miner is expected to maintain its performance superiority over different data factors \cite{e-NSP2}. Accordingly, here we empirically test the sensitivity of data factors discussed in Section \ref{datasets} on the \textit{sequence coverage} of EINSP and the  baselines. Table \ref{MS2} shows the sensitivity test on size $k=150$, where the adopted synthetic datasets are extended from the base dataset by tuning one boldface factor at a time. %Reliability

Table \ref{MS2} show the results of \textit{sequence coverage} of EINSP, which is always significantly higher than that of the baselines under different data factors. This is consistent with the results of the other comparisons. In addition, the \textit{sequence coverage} of EINSP is highly sensitive to factors \textit{C}, \textit{T} and \textit{N}. EINSP works better on the data with higher \textit{C}, higher \textit{T}, and lower \textit{N}. By increasing factor \textit{C}, more long-sized sequences are generated, and EINSP more likely covers the long-sized patterns, as shown in Section \ref{PatSize}. Moreover, a higher \textit{C} leads to a larger number of frequent elements being generated, and the corresponding diversity vector in EINSP contributes to a more diverse subset and makes EINSP more effective. 

For $k=150$, when \textit{C} increases from 6 to 14, the \textit{sequence coverage} of EINSP grows by 77.3\% and outperforms the baselines by an average of 55.8\% and a maximum of 73.6\%. Further, when factor \textit{T} grows, the dataset becomes denser and thus each pattern selected by EINSP covers more data sequences. For example, when \textit{T} increases from 4 to 12 for $k=150$, the \textit{sequence coverage} of EINSP increases by 85.4\% and is 17.0\% higher than that of k-SDPP, which is the second best method of all the baselines. 

In addition, the growth of factor \textit{DB} has a limited impact on all the methods because it does not change the distribution of a dataset. Lastly, EINSP achieves a lower \textit{sequence coverage} on those datasets with higher \textit{N}. This is due to the fact that (1) increasing factor \textit{N} produces more sparse data sequences, leading to a lower proportion of its sequences being covered by a small-scale subset; and (2) less frequent elements are available in a sparse dataset, making the diversity vector in EINSP less effective. In summary, the data factor sensitivity analysis further confirms the design of EINSP for more diverse NSP selection.

\section{Discussion}
As discussed in Section \ref{secintro}, NSA and NSP mining are often more informative for decision-making particularly for problems and applications with negative feedback such as in recommender systems, non-occurring behaviors and events such as in medical and health treatments, and risk, safety and security matters such as undeclared behaviors. However, the limited research and available algorithms on NSA prevent its widespread applications. 

The non-occurrence nature of negative sequences hide their great potential for applications. The various combinatorial issues further make NSA and NSP mining much more challenging than PSA and PSP mining, which has been widely explored in areas such as genomic analysis and pattern mining. A critical issue in existing sequence analysis is the underlying frequentist-based pattern selection on the downward closure property held between patterns. This inflicts significantly higher restriction on NSA than PSA due to the more sophisticated combinatorial scenarios of negative items, elements, itemsets, and patterns. 

This paper introduces a new way of discovering not only significant but also diverse and indirectly coupled NSPs based on the DPP-based NSP representations. Our work carries the study of pattern relation analysis \cite{cao2013combined} and coupling learning \cite{Coupling2015cao} into NSP mining to disclose the indirect couplings between NSP items, elements and patterns that do not co-occur. These cannot be achieved by the frequentist and downward closure property. 

Accordingly, EINSP opens a new direction, i.e., NSA with pattern selection by analyzing both explicit and implicit pattern relations and element relations, which can be applied to PSA and PSP mining \cite{cao2013combined}. More theoretical studies are required to explore comprehensive explicit and implicit element/pattern relations and structural relations in NSP pattern relation analysis. Examples are involving item constraint, element constraint, size constraint, format constraint, and combinatorial constraint \cite{WangC19,e-NSP2} into the graph-based NSP representations and NSA. Other opportunities lie in theoretical research on analyzing relations between elements and patterns in extremely sparse and diverse but large sequential data.

Last but not least, this work argues the importance of discovering \textit{actionable NSPs} that are significant, diverse, novel and informative by involving NSP statistics, diversity, and informativeness in terms of element and pattern couplings. Very limited research is available on \textit{pattern actionability} and \textit{actionable knowledge discovery}, which are increasingly essential in discovering \textit{actionable intelligence} for AI to inform decision-making. 

\section{Conclusion}  \label{secCon}

While \textit{negative sequence analysis} (NSA) has been rarely studied, it has played a strong role in discovering significant occurring and non-occurring entities, events and behaviors. Many significant theoretical and practical challenges surround NSA, e.g., a lack of theoretical foundation, complicated combinatorial scenarios, non-occurring relations, and extreme computational cost. 

This work represents a new direction in NSA to discover representative negative sequential patterns that are of high quality, diversity and informativeness, i.e., toward actionable NSP discovery. We propose a novel DPP-based NSP graph representation and the DPP-based EINSP models both explicit and implicit relations in positive and negative sequential elements and patterns in terms of the co-occurring and non-occurring probabilistic distributions over all possible subsets in the pattern cohort. EINSP selects a representative NSP  subset composed of high-quality and diverse NSP elements and patterns. Such actionable NSP discovery deserves further research with various potentials and challenges.

Our work is also new in proposing DPP-based NSP graph representation and NSP pattern relation analysis. Both theoretical and empirical analyses demonstrate the strong potential of EINSP in discovering more representative and informative patterns with wider coverage, larger diversity, and higher quality in terms of different data factors. The NSP graph representation and pattern relation analysis opens various opportunities for NSA and general sequence analysis.

% or


\iffalse
\appendices
\section{Proof of the First Zonklar Equation}
Appendix one text goes here.

\section{}
Appendix two text goes here.

\ifCLASSOPTIONcaptionsoff
  \newpage
\fi
\fi

\section*{Acknowledgment}
This work is partially sponsored by the Australian Research Council Discovery Grant DP190101079 and Future Fellowship Grant FT190100734. 



\bibliographystyle{IEEEtran}
\bibliography{myRef_abbr}%\bibliography{myRef}%\bibliography{IEEEabrv,../bib/paper}
%\begin{thebibliography}{1}

%\end{thebibliography}


%\end{IEEEbiography}


%\newpage
\begin{IEEEbiography}[{\includegraphics[width=1in,height=1.25in,clip,keepaspectratio]{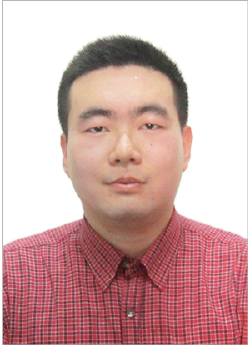}}]{Wei Wang} %\begin{IEEEbiography}[{\includegraphics[width=1in,height=1.25in,clip,keepaspectratio]{Fig/WeiWang}}]{Wei Wang}
received a PhD in analytics at the Advanced Analytics Institute, University of Technology Sydney, Australia. His research interests include data mining, machine learning, recommender systems, and software engineering.
\end{IEEEbiography}

\begin{IEEEbiography}[{\includegraphics[width=1in,height=1.25in,clip,keepaspectratio]{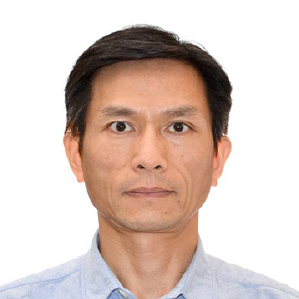}}]{Longbing Cao} %\begin{IEEEbiography}[{\includegraphics[width=1in,height=1.25in,clip,keepaspectratio]{Fig/Cao}}]{Longbing Cao}
is a Professor at the University of Technology Sydney and an ARC Future Fellow (Level 3). He has one PhD in Pattern Recognition and Intelligent Systems from the Chinese Academy of Sciences and another in Computing Science from UTS. His research interests include artificial intelligence, data science, knowledge discovery, machine learning, behavior informatics, complex intelligent systems, and their enterprise applications.
\end{IEEEbiography}
\iffalse
\begin{IEEEbiography}[{\includegraphics[width=1in,height=1.25in,clip,keepaspectratio]{Fig/IEEEbiography/kumar_photo.jpg}}]{Vipin Kumar} 
is a Regents Professor and the William Norris Endowed Chair at the University of Minnesota. His research interests span data mining, high-performance computing, and their applications in Climate/Ecosystems and health care. He is an ACM, IEEE, AAAS and SIAM fellow. He authored over 300 research articles, and has coedited or coauthored 10 books including two text books. He received numerous awards and chaired many conferences in the area of data mining, big data, and high performance computing.
\end{IEEEbiography}
\fi


%\vfill


\end{document}